\documentclass{article}
\usepackage{enumitem}
\usepackage{pifont}
 \usepackage[main, final]{neurips_2026}
\usepackage[utf8]{inputenc} % allow utf-8 input
\usepackage[T1]{fontenc}    % use 8-bit T1 fonts
\usepackage[colorlinks=true]{hyperref}       % hyperlinks
\usepackage{url}            % simple URL typesetting
\usepackage{booktabs}       % professional-quality tables
\usepackage{amsfonts}       % blackboard math symbols
\usepackage{nicefrac}       % compact symbols for 1/2, etc.
\usepackage{microtype}      % microtypography
\usepackage{xcolor}         % colors
\definecolor{best}{RGB}{0,128,0}      % dark green
\definecolor{second}{RGB}{0,102,204}  % blue
\usepackage[ruled, vlined]{algorithm2e}

\usepackage[colorlinks=true]{hyperref}

\definecolor{citecolor}{HTML}{0071bc}
\definecolor{linkcolor}{HTML}{c00000}
\definecolor{urlcolor}{HTML}{008000}

\hypersetup{
  citecolor=citecolor,
  linkcolor=linkcolor,
  urlcolor=urlcolor
}

\usepackage{algpseudocode}

\usepackage{amsmath,amssymb}

\usepackage{tikz}
\usetikzlibrary{arrows.meta, positioning, fit}

\usepackage{hyperref}

\newcommand{\ourmodel}{\textbf{Thermo-VL}}

\usepackage{amsmath}
\usepackage{amssymb}
\usepackage{mathtools}
\usepackage{amsthm}

\theoremstyle{plain}

\theoremstyle{definition}

\theoremstyle{remark}

\defcitealias{chen2024internvl}{IVL}
\defcitealias{Qwen-VL}{QwenVL}
\defcitealias{liu2024llavanext}{LlaVANext}
\defcitealias{deitke2025molmo1}{Molmo}

\title{\ourmodel{}: Extending Vision–Language Models\\ to Thermal Infrared Perception}

\author{%
\textbf{Rusiru Thushara} \quad
\textbf{Yasiru Ranasinghe} \quad
\textbf{Jay Paranjape} \quad
\textbf{Vishal M. Patel} \\\\
Department of Electrical \& Computer Engineering, \\ Johns Hopkins University, Baltimore, MD 21218 USA
}

\begin{document}

\maketitle

\begin{abstract}

Vision-language models (VLMs) often fail under low illumination because their visual grounding is learned predominantly from RGB imagery, whereas thermal infrared preserves complementary scene structure when visible cues degrade. We present \ourmodel{}, a wavelength-aware VLM that augments a frozen Molmo-7B backbone with a trainable thermal encoder and a text-guided dual-attention fusion module. Given aligned RGB tokens, thermal tokens, and prompt embeddings, the fusion module conditions thermal features on both language and RGB context, then injects a gated residual into the frozen RGB stream so thermal evidence can be incorporated without disrupting Molmo's pretrained RGB--language interface. We train the model with the standard language-modeling objective together with auxiliary alignment and regularization losses that improve cross-modal grounding and reduce over-reliance on RGB. We also introduce a pixel-aligned RGB--thermal instruction-tuning dataset and Thermo-VL-Bench, a manually screened RGB--thermal VQA benchmark for low-light and cross-spectrum reasoning. Experiments show strong gains on challenging thermal-only and RGB+thermal reasoning tasks, highlighting the value of prompt-conditioned multispectral fusion. Our dataset and code are publicly available at: \href{https://thusharakart.github.io/Thermo-VL}{thusharakart.github.io/Thermo-VL}

\end{abstract}

\section{Introduction}

Recent advances in vision-language models (VLMs) have dramatically improved open-ended image understanding, captioning, and visual question answering by combining large-scale multimodal pretraining with instruction tuning~\cite{deitke2025molmo, liu2023visual,huang2025visual}. However, most current VLMs are trained almost exclusively on visible-spectrum RGB imagery and therefore inherit the failure modes of RGB sensing. Under low illumination, glare, haze, smoke, or other adverse conditions, RGB observations can lose critical scene evidence even when reliable reasoning is most needed.

Thermal infrared provides complementary information in precisely these settings. Because it captures radiometric structure rather than reflected visible light, thermal imagery can preserve salient cues such as humans, animals, and warm vehicles when visible features degrade~\cite{zhao2024removal}. This mismatch makes RGB-only VLMs brittle in safety-critical scenarios such as nighttime driving, search-and-rescue, and low-visibility inspection.

A common workaround is to translate infrared imagery into pseudo-RGB and then apply an RGB-trained VLM~\cite{shyam2024lightweight}. However, translation-based pipelines introduce a generative bottleneck, may hallucinate visible-style details, and can suppress modality-specific information that is not naturally represented in RGB space. A more principled alternative is to integrate thermal observations directly inside the VLM. The challenge is to do so without destabilizing the pretrained RGB--language interface or requiring expensive retraining of the backbone. Progress is further limited by the lack of language-supervised RGB--thermal data: widely used paired datasets such as KAIST, FLIR-ADAS, and OSU/OTCBVS largely target detection or tracking and provide little to no VQA-style supervision.

\begin{figure}[t]
    \centering
    \includegraphics[width=\textwidth]{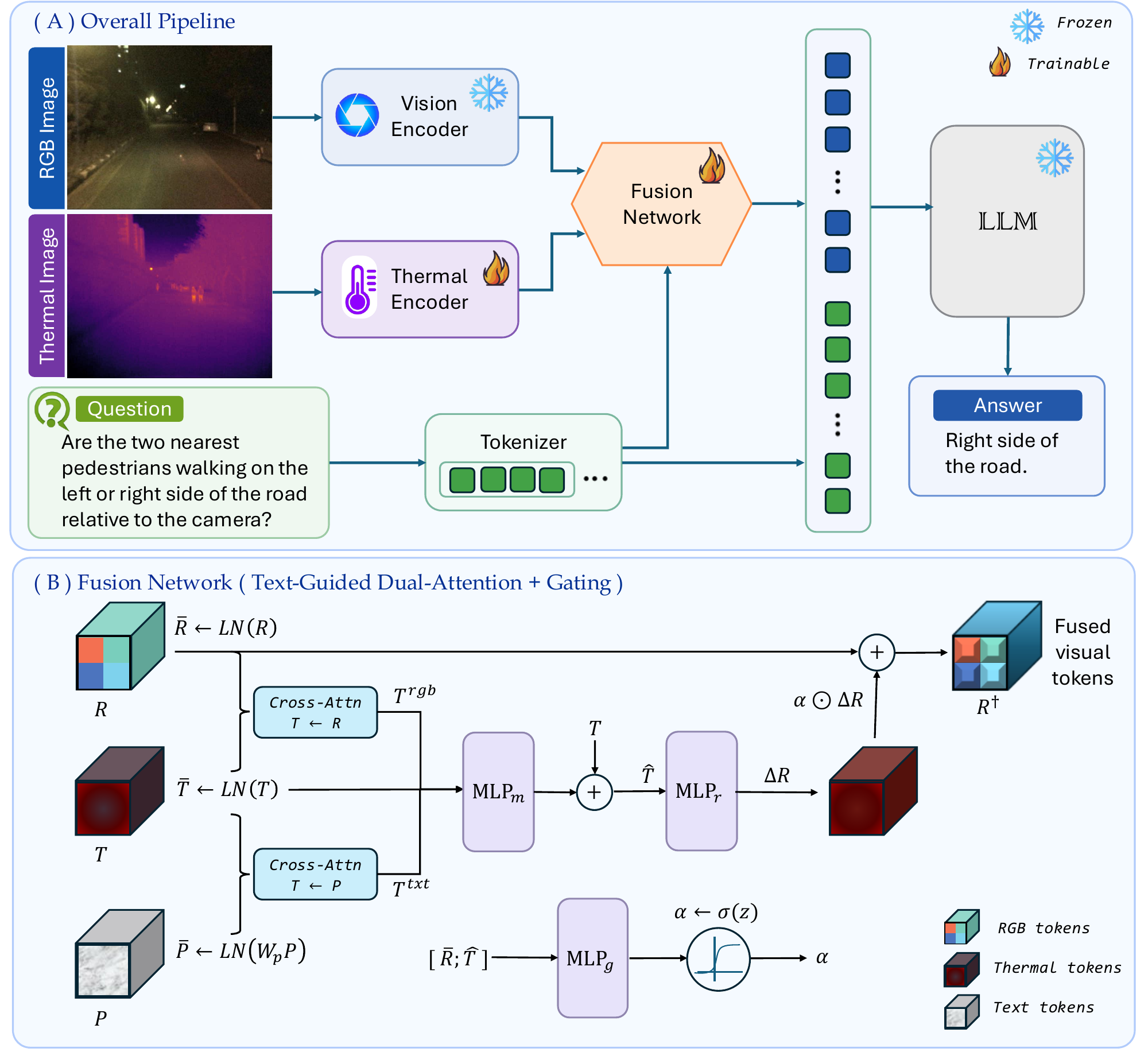}
    \caption{\ourmodel{} overview. (A) Overall pipeline: a frozen RGB vision tower and a finetuned thermal encoder produce aligned patch tokens from paired RGB and thermal inputs. Prompt embeddings guide a text-aware fusion module, and the frozen language model processes the fused visual tokens to generate the final output. (B) Text-guided dual-attention fusion block: thermal tokens attend to both prompt embeddings and RGB tokens; the refined thermal representation predicts residual and token-wise gates, and the gated residual is injected into RGB tokens to form fused visual tokens \(\mathbf{R}^{\dagger}\).} 
    \label{fig:model-architecture}
\end{figure}

In this work, we present \ourmodel{}, a wavelength-aware VLM built on Molmo-7B-D~\cite{deitke2025molmo}. Our model combines a frozen RGB vision tower with a partially trainable thermal encoder and a text-guided dual-attention fusion module. Given aligned RGB tokens, thermal tokens, and prompt embeddings, the fusion module first refines thermal features using both the question and RGB context, then injects the resulting information into the frozen RGB stream through a token-wise gated residual. This design makes fusion question-aware, preserves compatibility with Molmo's pretrained language stack, and supports parameter-efficient adaptation by updating only the last \(N_\ell\) blocks of the thermal encoder together with the fusion parameters. The overall architecture is shown in Fig.~\ref{fig:model-architecture}(A), and the detailed fusion block is shown in Fig.~\ref{fig:model-architecture}(B).

To support instruction tuning and standardized evaluation in this setting, we also introduce a unified RGB--thermal data suite. We construct a pixel-aligned RGB--thermal instruction-tuning dataset with question--answer pairs spanning scene understanding, counting, spatial relations, and safety-relevant cues. We further introduce \textbf{Thermo-VL-Bench}, a human-verified RGB--thermal VQA benchmark designed for low-light and cross-spectrum reasoning.

% Experiments show that explicit, question-aware RGB--thermal fusion yields consistent gains over strong VLM baselines, especially when both modalities are available. These results highlight that thermal cues are not merely redundant to RGB, but complementary when integrated at the token level.

Experiments on Thermo-VL-Bench and RGB-Th Bench show that explicit prompt-conditioned RGB--thermal fusion is especially helpful when both modalities are available. Relative to unmodified VLM baselines evaluated with paired inputs, \ourmodel{} preserves RGB-only performance while improving thermal-only and RGB+thermal settings. These results suggest that thermal cues can complement RGB when fused inside the model rather than appended at the input level.

\paragraph{Contributions.}
Our main contributions are summarized as follows:
\vspace{-0.2em}
\begin{itemize}[leftmargin=1.2em,itemsep=3pt,topsep=0pt,parsep=1pt]
    \item \textbf{Wavelength-aware VLM with question-aware fusion:} We introduce \ourmodel{}, a Molmo-based RGB--thermal VLM that integrates paired RGB and thermal inputs through a text-guided dual-attention fusion module with token-wise gated residual injection, while keeping the RGB tower and language model frozen.
    \item \textbf{Parameter-efficient multispectral adaptation:} We fine-tune only the last \(N_\ell\) blocks of the thermal encoder together with the thermal LayerNorm and fusion module, and regularize training with token-level RGB--thermal alignment, visual--thermal--text contrastive supervision, gate-entropy regularization, and block-wise RGB masking.
    \item \textbf{Unified RGB--thermal language supervision:} We introduce a pixel-aligned RGB--thermal instruction-tuning dataset and Thermo-VL-Bench, a manually screened binary benchmark built from real aligned image pairs, to support training and initial evaluation in low-light cross-spectrum settings.
\end{itemize}
\vspace{-0.3em}

\section{Related Work}

Vision-language models have advanced rapidly, enabling open-ended captioning and visual question answering (VQA) on RGB images through instruction tuning on large-scale datasets such as PixMo~\cite{deitke2025molmo}. Models like Molmo-7B, which \ourmodel{} directly extends, fuse a frozen ViT vision tower with a language model connected via a projector, trained on detailed captions and synthetic QA pairs~\cite{deitke2025molmo}. These models perform well in natural lighting but typically overlook wavelength-specific inputs, limiting performance in low-light or adverse weather.

\noindent\textbf{Thermal and Infrared Language Understanding.}
Thermal and infrared language understanding remains underexplored, with limited prior work on direct thermal input processing for captioning and VQA~\cite{jiang2024infraredllava}. Existing approaches often translate thermal images to pseudo-RGB before applying RGB-trained VLMs, or use language models to mine captions from infrared imagery without native cross-spectrum fusion~\cite{jiang2024infraredllava}. These strategies do not directly align thermal observations with language and therefore remain limited in settings where visible-spectrum cues degrade. 

\noindent\textbf{RGB--Thermal Fusion Mechanisms.}
RGB--thermal fusion has been widely studied in detection and segmentation, where attention and gating modules dynamically combine visible and thermal cues~\cite{el2023enhanced, sun2019rtfnet, xie2023illumination, wu2025atdetr, li2025multimodal, rathinam2024hybrid}. Most of this literature focuses on closed-set perception tasks such as object detection or pedestrian recognition. In contrast, our setting requires open-ended language generation and question answering within a pretrained VLM, where fusion must preserve the existing RGB--language interface. 
%Our setting differs from RGB--thermal detection and segmentation in that the fused representation must remain compatible with a pretrained language stack rather than a task-specific prediction head.

\noindent\textbf{Visible-to-Thermal Translation.}
Visible-to-thermal translation has emerged as a useful tool for data augmentation and cross-spectrum synthesis~\cite{paranjape2025f, xiao2025thermalgen, o2015introduction, berg2018generating, li2023feasibility}. While such methods can expand training data, they are not a substitute for native multispectral reasoning because the downstream model still needs to ground thermal evidence in language. In this work, we use synthetic thermal generation as augmentation only, while the proposed model reasons directly over paired RGB and thermal inputs.

\noindent\textbf{RGB--Thermal Datasets.}
Public paired datasets such as LLVIP~\cite{jia2021llvip}, KAIST~\cite{hwang2015multispectral}, FLIR~\cite{flir_adas}, and OSU/OTCBVS~\cite{otcbvs_osu_thermal} provide valuable cross-spectrum imagery, but most were created for detection, tracking, or recognition rather than language-supervised reasoning. Existing RGB--thermal VQA resources remain small and primarily evaluation-oriented. This lack of standardized training-scale language annotations motivates our dataset construction pipeline and benchmark design. Details are provided in Sec.~\ref{sec:data}. 

\section{Method}

\subsection{Overview}

Our model, \ourmodel{}, builds on the Molmo-7B-D backbone and augments its frozen RGB--language interface with a partially trainable thermal branch and a lightweight fusion module. Given a paired RGB image \(I^{\mathrm{rgb}}\), thermal image \(I^{\mathrm{th}}\), and prompt \(p\), the model first encodes both modalities into spatially aligned patch tokens. The fusion module then uses the prompt and RGB context to refine thermal tokens and injects the resulting information into the frozen RGB stream through a gated residual. The fused visual tokens are finally passed to the frozen language model for autoregressive generation. Figure~\ref{fig:model-architecture}(A) summarizes the overall pipeline.

\subsection{Visual Encoders and Input Preprocessing}

\ourmodel{} uses two aligned visual streams built on the Molmo-7B-D vision architecture. The RGB branch uses Molmo's pretrained vision tower \(\phi_R\) and remains frozen throughout training. The thermal branch uses a ViT-L/14 encoder \(\phi_T\) initialized from the MAE-pretrained checkpoint described in Appendix Sec.~\ref{sec:mae_encoder}. Because the RGB and thermal towers share the same patch embedding geometry and hidden dimension, both branches produce \(N\) spatially aligned patch tokens in \(\mathbb{R}^{N\times d}\), enabling one-to-one token-level fusion without interpolation or an additional projection layer.

Molmo's vision encoder expects 3-channel inputs. Therefore, each single-channel thermal image is replicated across three channels before applying the standard Molmo image preprocessing pipeline. This preserves compatibility with the pretrained vision stack while keeping the thermal signal radiometrically single-channel. Thermal images shown in the figures are heatmap-colorized for visualization only and are not used as model inputs.

During instruction tuning, we freeze the RGB encoder and update only the last \(N_\ell\) transformer blocks of the thermal encoder together with the thermal LayerNorm \(\mathrm{LN}_T\) and the fusion module. In all experiments, we set \(N_\ell=4\). This preserves low- and mid-level infrared structure learned during MAE pretraining while adapting higher-level thermal semantics to the downstream language objective.

% \begin{figure}[t]
%     \centering
%     \includegraphics[width=\linewidth]{figures/bidirectional-fusion-gating.pdf}
%     \caption{Text-guided dual-attention fusion block. Prompt tokens are projected into the visual space and used, together with RGB tokens, to refine thermal tokens through two attention paths. The refined thermal representation predicts a residual and token-wise gates, and the gated residual is added to RGB tokens before they are passed to the frozen LLM.}
%     \label{fig:patch_gating_network}
% \end{figure}

\begin{figure}[!t]
    \centering
    \includegraphics[width=\linewidth]{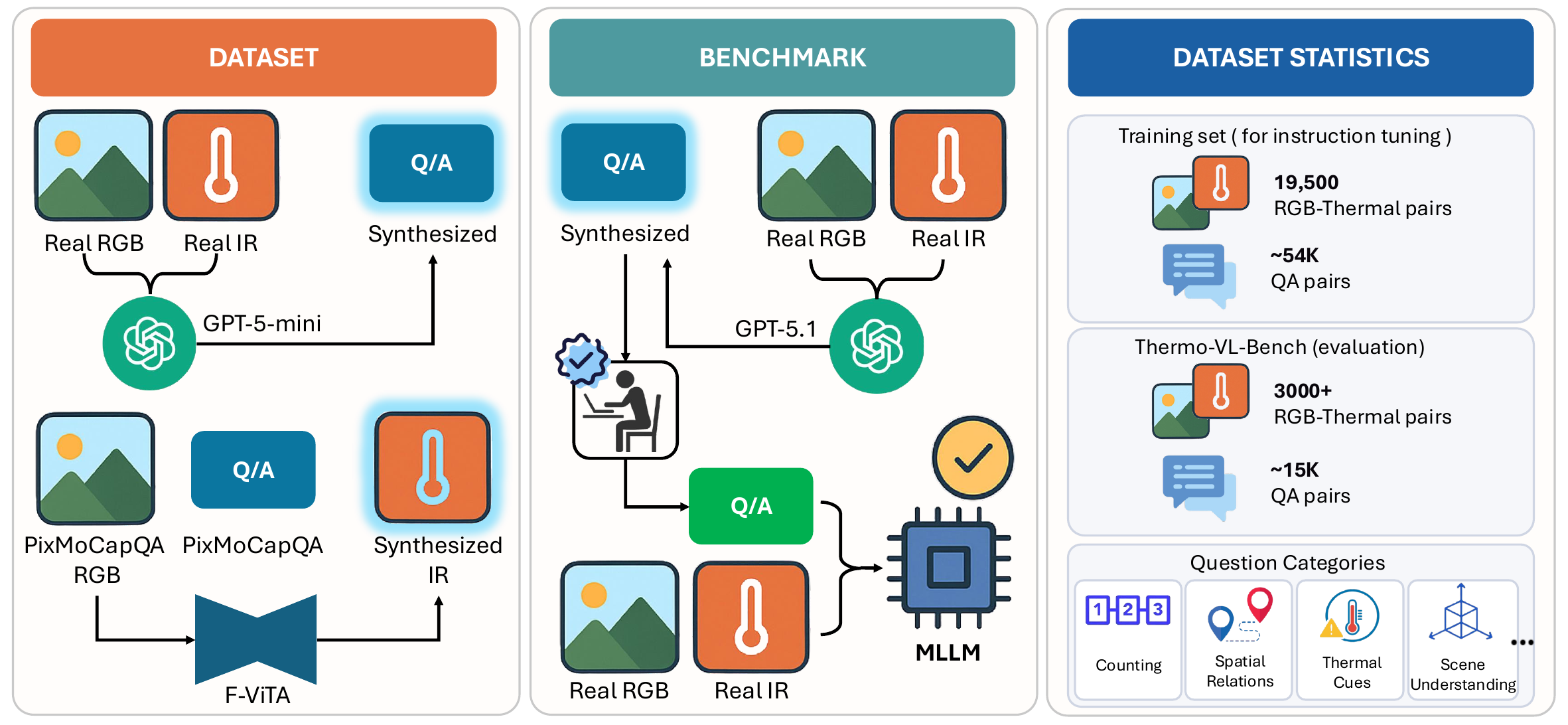}
    \caption{Construction of the RGB--thermal training data and Thermo-VL-Bench. For training, we pair real RGB images with real thermal images or synthetic thermal counterparts generated by F-ViTA and obtain QA supervision using GPT-5-mini or inherited PixMo-cap-qa annotations. For evaluation, candidate QA pairs are generated using GPT-5.1 from real aligned pairs and then manually verified.}
    \label{fig:dataset-benchmark}
\end{figure}

\subsection{Text-Guided Dual-Attention Fusion}
As shown in Fig.~\ref{fig:model-architecture}(B), we introduce a text-guided dual-attention fusion block that refines thermal evidence before injecting it into the frozen RGB token stream. Let \(\mathbf{R}\in\mathbb{R}^{N\times d}\), \(\mathbf{T}\in\mathbb{R}^{N\times d}\), and \(\mathbf{P}\in\mathbb{R}^{L\times d_p}\) denote RGB tokens, thermal tokens, and prompt embeddings. We normalize inputs and project prompts into the visual space as \(\bar{\mathbf{R}}=\operatorname{LN}(\mathbf{R})\), \(\bar{\mathbf{T}}=\operatorname{LN}(\mathbf{T})\), and \(\bar{\mathbf{P}}=\operatorname{LN}(W_P\mathbf{P})\), where \(W_P\) maps prompts to dimension \(d\).

We form two complementary thermal-context paths: a prompt-conditioned path \(\mathbf{T}^{\mathrm{txt}}=\operatorname{MHA}(\bar{\mathbf{T}},\bar{\mathbf{P}},\bar{\mathbf{P}};M_P)\) and an RGB-conditioned path \(\mathbf{T}^{\mathrm{rgb}}=\operatorname{MHA}(\bar{\mathbf{T}},\bar{\mathbf{R}},\bar{\mathbf{R}})\). These are merged with the original thermal features to obtain
\[
\hat{\mathbf{T}}=
\operatorname{LN}\!\Bigl(
\mathbf{T}+
\operatorname{MLP}_m\bigl([\bar{\mathbf{T}};\mathbf{T}^{\mathrm{txt}};\mathbf{T}^{\mathrm{rgb}}]\bigr)
\Bigr).
\]

Rather than replacing RGB tokens, we predict a residual \(\Delta\mathbf{R}=\operatorname{MLP}_r(\operatorname{LN}(\hat{\mathbf{T}}))\) and modulate it with token-wise gates \(\boldsymbol{\alpha}=\sigma(\operatorname{MLP}_g(\operatorname{LN}([\bar{\mathbf{R}};\hat{\mathbf{T}}])))\), where \(\boldsymbol{\alpha}\in(0,1)^{N\times 1}\). The final fused tokens are
\[
\mathbf{R}^{\dagger}=\mathbf{R}+\boldsymbol{\alpha}\odot\Delta\mathbf{R}.
\]

This design makes fusion query-conditional rather than a fixed preprocessing step. Thermal tokens interact with both prompt and RGB context, emphasizing task-relevant temperature cues. The gated residual preserves the frozen RGB interface of the pretrained VLM, stabilizing language grounding while enabling targeted thermal injection. Alignment and contrastive regularization organize RGB, thermal, and text features in a shared space, while gate entropy and block-wise RGB masking discourage over-reliance on the visible branch and prevent modality collapse, yielding a parameter-efficient architecture for robust cross-spectrum reasoning.

\subsection{Training Objective}

We train \ourmodel{} while freezing the RGB vision encoder \(\phi_R\) and the language model \(g_{\mathrm{LM}}\). The trainable parameters consist of the last \(N_\ell\) blocks of the thermal encoder \(\phi_T\), the thermal LayerNorm \(\mathrm{LN}_T\), and the fusion module \(\mathcal{F}_\theta\).

\paragraph{Block-wise RGB masking.}
To reduce over-reliance on the visible-spectrum branch, we randomly mask contiguous square regions whose union covers 10\% of the RGB image area during training, while leaving the thermal input unchanged. Unlike token-level patch dropout, this augmentation operates in image space and acts as a coarse corruption augmentation for the visible branch. The masking is disabled at inference time.

\paragraph{Objective.}
Given fused tokens \(\mathbf{R}^{\dagger}=\mathcal{F}_\theta(\mathbf{R},\mathbf{T},\mathbf{P})\), the frozen language model predicts \(\hat{y}=g_{\mathrm{LM}}(\mathbf{R}^{\dagger},\mathbf{P})\), and we optimize
\[
\mathcal{L}
=
\mathcal{L}_{\mathrm{LM}}
+\lambda_{\mathrm{align}}\mathcal{L}_{\mathrm{align}}
+\lambda_{\mathrm{contr}}\mathcal{L}_{\mathrm{contr}}
+\lambda_{\mathrm{gate}}\mathcal{L}_{\mathrm{gate}}.
\]

The main task term is the autoregressive cross-entropy loss \(\mathcal{L}_{\mathrm{LM}}=\operatorname{CE}(\hat{y},y^\star)\). We enforce token-level RGB--thermal consistency via \(\mathcal{L}_{\mathrm{align}}=\frac{1}{N}\sum_{n=1}^{N}\lVert \mathbf{R}_n-\mathbf{T}_n\rVert_2^2\).

For global semantic consistency, we pool each stream as \(\mathbf{p}_R=\operatorname{Pool}(\mathbf{R}^{\dagger})\), \(\mathbf{p}_T=\operatorname{Pool}(\mathbf{T})\), and \(\mathbf{p}_P=\operatorname{Pool}_{M_P}(\mathbf{P})\), and define
\[
\mathcal{L}_{\mathrm{contr}}
=
\frac{1}{2}\Bigl[
\operatorname{NCE}(\mathbf{p}_R,\mathbf{p}_T;\tau)
+
\operatorname{NCE}(\mathbf{p}_R,\mathbf{p}_P;\tau)
\Bigr],
\]
where \(\operatorname{NCE}\) is InfoNCE with temperature \(\tau\) and in-batch negatives.

To avoid gate saturation, we maximize entropy \(H(\alpha_n)=-\alpha_n\log\alpha_n-(1-\alpha_n)\log(1-\alpha_n)\) by minimizing
\[
\mathcal{L}_{\mathrm{gate}}
=
-\frac{1}{N}\sum_{n=1}^{N} H(\alpha_n).
\]

\paragraph{Optimization.}
We use AdamW with two parameter groups. Group-T contains the last \(N_\ell\) blocks of the thermal encoder and is updated with learning rate \(\eta_{\mathrm{therm}}\). Group-F contains the fusion parameters and \(\mathrm{LN}_T\) and is updated with learning rate \(\eta_{\mathrm{fusion}}\). This separation allows conservative adaptation of the pretrained thermal encoder while enabling faster learning in the newly initialized fusion parameters.

\section{Datasets and Benchmark}
\label{sec:data}

To support multispectral instruction tuning and evaluation, we construct (i) a training set of aligned RGB--thermal question--answer pairs and (ii) \textbf{Thermo-VL-Bench}, a human-verified evaluation benchmark. The training data combines real paired RGB--thermal imagery from public datasets with synthetic thermal counterparts generated for RGB-only images from PixMo-cap-qa~\cite{deitke2025molmo}. The benchmark uses only real, pixel-aligned RGB--thermal image pairs. Figure~\ref{fig:dataset-benchmark} summarizes the overall data pipeline.

% \subsection{Training Data Construction}
\paragraph{Training Data Construction}
We collect aligned RGB--thermal image pairs from public datasets including KAIST~\cite{hwang2015multispectral}, FLIR-ADAS~\cite{flir_adas}, and OSU/OTCBVS~\cite{otcbvs_osu_thermal}. For real paired examples without language annotations, we use OpenAI GPT-5-mini to generate question--answer pairs grounded in the RGB image, the thermal image, or both modalities jointly. The prompts are designed to cover scene understanding, counting, spatial relations, and safety-relevant cues that are often easier to identify in thermal imagery. This process yields instruction-tuning samples for RGB-only, thermal-only, and RGB+thermal reasoning from the same aligned scene.

% \subsection{Synthetic Thermal Augmentation}
\paragraph{Synthetic Thermal Augmentation.}
To increase data scale and linguistic diversity, we extend RGB-only images from PixMo-cap-qa~\cite{deitke2025molmo} with synthetic thermal counterparts. We first filter PixMo-cap-qa for natural scenes using CLIP~\cite{radford2021learning}, then use F-ViTA~\cite{paranjape2025f} to generate aligned thermal images. We retain the original QA supervision from PixMo-cap-qa and pair it with the aligned RGB--thermal inputs. These synthetic examples broaden the range of training scenes and question types while preserving the paired structure required for multispectral instruction tuning.

% \subsection{Thermo-VL-Bench}

% Given the limited availability of standardized RGB--thermal VQA benchmarks, we introduce \textbf{Thermo-VL-Bench}, a human-verified evaluation set built from real, pixel-aligned RGB--thermal image pairs. We first use OpenAI GPT-5.1 to generate candidate yes/no questions under three settings: RGB-only, thermal-only, and RGB+thermal. All benchmark QA pairs are then manually reviewed by the authors. During this process, approximately 20\% of candidate QA pairs are removed for being ambiguous, inconsistent with the intended modality, or otherwise unsuitable. Accepted items are kept verbatim after screening so that the benchmark follows a fixed generate-and-screen protocol rather than post-hoc editorial rewriting. This yields a controlled benchmark for measuring modality-specific reasoning and the gains from RGB--thermal fusion under low-light and cross-spectrum conditions. We use binary questions in the benchmark because they make modality-specific answerability easier to verify and allow unambiguous automatic scoring across models.

\paragraph{Thermo-VL-Bench.}
Given the limited availability of standardized RGB--thermal language benchmarks, we introduce \textbf{Thermo-VL-Bench}, a manually screened binary evaluation set built from real, pixel-aligned RGB--thermal image pairs. We first use OpenAI GPT-5.1 to propose candidate yes/no questions under three conditions: RGB-only, thermal-only, and RGB+thermal. We then manually screen each candidate for (i) answerability from the designated modality, (ii) consistency with the image pair, and (iii) absence of ambiguity or unintended reliance on the other modality. Approximately 20\% of the candidate items are discarded during this process. Accepted items are kept verbatim after screening, so that the benchmark follows a fixed generate-and-screen protocol rather than post hoc editorial rewriting. We view Thermo-VL-Bench as an initial controlled evaluation resource rather than a fully human-authored benchmark.

% \subsection{Dataset Statistics}
\paragraph{Dataset Statistics.}
Our training set contains 19,500 pixel-aligned RGB--thermal image pairs and approximately 54K QA pairs. The questions cover a range of reasoning categories, including counting, spatial relations, safety-related cues visible in thermal imagery, and detailed scene understanding. Thermo-VL-Bench contains 3,148 RGB--thermal image pairs and approximately 15K QA pairs across RGB-only, thermal-only, and RGB+thermal settings. The training and evaluation splits use disjoint image pairs. Table~\ref{tab:rgbt_dataset_comparison} compares our data with prior RGB--thermal benchmarks and highlights that our dataset supports both large-scale instruction tuning and standardized evaluation. Representative training examples and additional data visualizations are provided in Appendix Fig.~\ref{fig:train_data_sample} and Fig.~\ref{fig:more_train_data_samples}.

\begin{figure*}[!t]
    \centering
    \includegraphics[width=\textwidth]{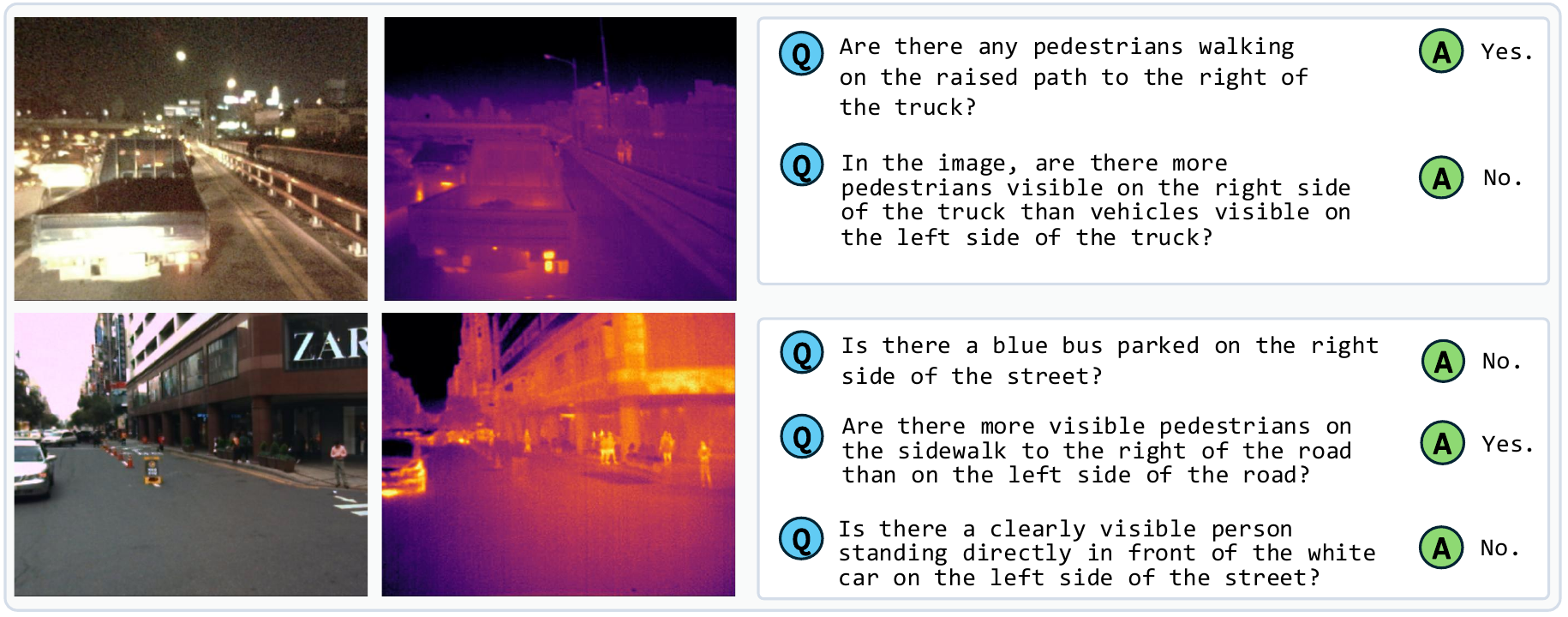}
    \caption{
        Representative samples from our evaluation benchmark. 
        Each sample includes an RGB image (left), a thermal image (right), and corresponding question--answer pairs.
        Thermal images are displayed using heatmap colorization to enhance visual interpretability.
    }
    \label{fig:eval_data_sample}
\end{figure*}

\begin{table*}[t]
\centering
\caption{
Comparison of existing RGB--thermal (and related) datasets and our proposed RGB--thermal VQA dataset.
Prior RGB--thermal datasets (e.g., KAIST, FLIR-ADAS, OSU, NIRSCENE) do not provide question--answer supervision.
RGB-Th-Bench offers an evaluation-only VQA benchmark, whereas our dataset supports both training ($\sim$54K QA pairs)
and evaluation ($\sim$15K QA pairs).
}
\setlength{\tabcolsep}{4pt}
\tiny
\resizebox{\columnwidth}{!}{%
\begin{tabular}{lcccccc}
\toprule
\textbf{Dataset} & \textbf{Modality} & \textbf{\#Images} & \textbf{Task /} & \textbf{\#Train} & \textbf{\#Eval} & \textbf{VQA} \\
                 &                   &                   & \textbf{Benchmark} & \textbf{QA pairs} & \textbf{QA pairs} & \textbf{Support} \\
\midrule
KAIST~\cite{hwang2015multispectral} & RGB + Th & $\sim$95K & Detection / Tracking & \ding{55} & \ding{55} & \ding{55} \\
FLIR-ADAS~\cite{flir_adas} & RGB + Th & $\sim$10K & Detection & \ding{55} & \ding{55} & \ding{55} \\
OSU (OTCBVS)~\cite{otcbvs_osu_thermal} & RGB + Th & 284 & Detection / Tracking & \ding{55} & \ding{55} & \ding{55} \\
NIRSCENE~\cite{epfl_rgb_nir_scene} & RGB + NIR & 477 & Classification / Recognition & \ding{55} & \ding{55} & \ding{55} \\
RGB-Th-Bench~\cite{moshtaghi2025rgb} & RGB + Th & 29 & VQA (evaluation only) & \ding{55} & 1624 & \ding{51} \\
% \midrule
\textbf{Thermo-VL-Bench (Ours)} & RGB + Th & \textbf{22648} & \textbf{VQA (train + eval)} & \textbf{$\sim$54K} & \textbf{$\sim$15K} & \textbf{\ding{51}} \\
\bottomrule
\end{tabular}%
}
\label{tab:rgbt_dataset_comparison}
\end{table*}

% \begin{table*}[h]
% \centering
% \caption{\textbf{Thermo-VL-Bench results.} Accuracy (higher is better) under RGB-only, thermal-only, and RGB+thermal inputs. \textit{Overall} is the QA-count-weighted average across the three subsets. Our method achieves the best overall score and benefits most from RGB and thermal fusion.}
% \label{tab:rgb_thermal_results}
% \resizebox{\textwidth}{!}{%
% \begin{tabular}{lcccccc}
% \toprule
% Model & LLM backbone & Vision backbone & RGB & IR & RGB+IR & Overall \\
% \midrule
% \texttt{InternVL 2.5-8B}~(\citetalias{chen2024internvl}) & \texttt{InternLM2.5-7B} & \texttt{InternViT-300M-v2.5} & 76.15 & 77.61 & 77.18 & 76.96 \\
% % \texttt{Qwen2-VL-7B-Instruct} & \texttt{Qwen2-7B} & \texttt{QwenViT} & 89.80 & 89.78 & 87.06 & 89.07 \\
% \texttt{Qwen2.5-VL-7B-Instruct}~(\citetalias{Qwen-VL}) & \texttt{Qwen2-7B} & \texttt{QwenViT} & 83.02 & 86.31 & 76.25 & 82.45 \\
% \texttt{LLaVA-Next-Llama3-8B}~(\citetalias{liu2024llavanext}) & \texttt{Llama-3-8B} & \texttt{CLIP ViT-L/14} & 89.97 & 78.67 & 85.89 & 84.73\\
% \texttt{Molmo-7B-D-0924}~(\citetalias{deitke2025molmo1}) & \texttt{Qwen2-7B} & \texttt{CLIP ViT-L/14} & 88.67 & 81.20 & 75.79 & 82.51 \\
% \texttt{\ourmodel(\textbf{Ours})} & \texttt{Qwen2-7B} & \texttt{CLIP-ViT-L/14} & 88.67 & 82.23 & 88.96 & 86.37 \\

% \bottomrule
% \end{tabular}%
% }
% \end{table*}

\begin{table*}[h]
\centering
\caption{\textbf{Thermo-VL-Bench results.} Accuracy (higher is better) under RGB-only, thermal-only, and RGB+thermal inputs. \textit{Overall} is the QA-count-weighted average across the three subsets. Best in green, second in blue.}
\label{tab:rgb_thermal_results}
\resizebox{\textwidth}{!}{%
\begin{tabular}{lcccccc}
\toprule
Model & LLM backbone & Vision backbone & RGB & IR & RGB+IR & Overall \\
\midrule
\texttt{InternVL 2.5-8B}~(\citetalias{chen2024internvl}) & \texttt{InternLM2.5-7B} & \texttt{InternViT-300M-v2.5} & 76.15 & 77.61 & 77.18 & 76.96 \\
\texttt{Qwen2.5-VL-7B-Instruct}~(\citetalias{Qwen-VL}) & \texttt{Qwen2-7B} & \texttt{QwenViT} & 83.02 & \textcolor{second}{86.31} & 76.25 & 82.45 \\
\texttt{LLaVA-Next-Llama3-8B}~(\citetalias{liu2024llavanext}) & \texttt{Llama-3-8B} & \texttt{CLIP ViT-L/14} & \textcolor{best}{89.97} & 78.67 & \textcolor{second}{85.89} & \textcolor{second}{84.73}\\
\texttt{Molmo-7B-D-0924}~(\citetalias{deitke2025molmo1}) & \texttt{Qwen2-7B} & \texttt{CLIP ViT-L/14} & \textcolor{second}{88.67} & 81.20 & 75.79 & 82.51 \\
\texttt{\ourmodel(\textbf{Ours})} & \texttt{Qwen2-7B} & \texttt{CLIP-ViT-L/14} & \textcolor{second}{88.67} & \textcolor{best}{82.23} & \textcolor{best}{88.96} & \textcolor{best}{86.37} \\
\bottomrule
\end{tabular}%
}
\end{table*}

% \begin{table*}[!h]
% \centering
% \caption{\textbf{RGB-Th Bench results}~\cite{moshtaghi2025rgb}. Performance (higher is better) under RGB-Txt and RGB-Th-Txt; \textit{Overall} averages across settings. Our method performs best overall, especially for RGB-Th-Txt.}
% \label{tab:rgb_th_eval_results}
% \resizebox{\textwidth}{!}{%
% \begin{tabular}{lccccc}
% \toprule
% Model & LLM backbone & Vision backbone & RGB-Txt & RGB-Th-Txt & Overall \\
% \midrule
% \texttt{InternVL 2.5-8B}~(\citetalias{chen2024internvl}) & \texttt{InternLM2.5-7B} & \texttt{InternViT-300M-v2.5} & 79.68 & 63.31 & 71.50 \\ 
% \texttt{Qwen2.5-VL-7B-Instruct}~(\citetalias{Qwen-VL}) & \texttt{Qwen2.5-7B} & \texttt{QwenViT} & 75.62 & 67.26 & 71.44 \\ %
% \texttt{LLaVA-Next-Mistral-7.6B}~(\citetalias{liu2024llavanext})  & \texttt{Mistral-7B} & \texttt{CLIP ViT-L/14} & 77.83 & 52.82 & 65.33 \\  
% \texttt{LLaVA-Next-Llama3-8B}~(\citetalias{liu2024llavanext})  & \texttt{Llama-3 8B} & \texttt{CLIP ViT-L/14}  & 77.22 & 58.63 & 67.92 \\  % [web:26][web:30]
% \texttt{Molmo-7B-D-0924}~(\citetalias{deitke2025molmo1})  & \texttt{Qwen2-7B} & \texttt{CLIP ViT-L/14} & 77.25 & 60.84 & 69.05 \\
% \texttt{\ourmodel(\textbf{Ours})}   & \texttt{Qwen2-7B} & \texttt{CLIP-ViT-L/14} & 77.25 & 69.15 & 73.20 \\
% \bottomrule
% \end{tabular}%
% }
% \end{table*}

\begin{table*}[!h]
\centering
\caption{\textbf{RGB-Th Bench results}~\cite{moshtaghi2025rgb}. Performance (higher is better) under RGB-Txt and RGB-Th-Txt; \textit{Overall} averages across settings. Best in green, second in blue.}
\label{tab:rgb_th_eval_results}
\resizebox{\textwidth}{!}{%
\begin{tabular}{lccccc}
\toprule
Model & LLM backbone & Vision backbone & RGB-Txt & RGB-Th-Txt & Overall \\
\midrule
\texttt{InternVL 2.5-8B}~(\citetalias{chen2024internvl}) & \texttt{InternLM2.5-7B} & \texttt{InternViT-300M-v2.5} & \textcolor{best}{79.68} & 63.31 & \textcolor{second}{71.50} \\ 
\texttt{Qwen2.5-VL-7B-Instruct}~(\citetalias{Qwen-VL}) & \texttt{Qwen2.5-7B} & \texttt{QwenViT} & 75.62 & \textcolor{second}{67.26} & 71.44 \\ 
\texttt{LLaVA-Next-Mistral-7.6B}~(\citetalias{liu2024llavanext}) & \texttt{Mistral-7B} & \texttt{CLIP ViT-L/14} & \textcolor{second}{77.83} & 52.82 & 65.33 \\  
\texttt{LLaVA-Next-Llama3-8B}~(\citetalias{liu2024llavanext}) & \texttt{Llama-3 8B} & \texttt{CLIP ViT-L/14}  & 77.22 & 58.63 & 67.92 \\  
\texttt{Molmo-7B-D-0924}~(\citetalias{deitke2025molmo1}) & \texttt{Qwen2-7B} & \texttt{CLIP ViT-L/14} & 77.25 & 60.84 & 69.05 \\
\texttt{\ourmodel(\textbf{Ours})} & \texttt{Qwen2-7B} & \texttt{CLIP-ViT-L/14} & 77.25 & \textcolor{best}{69.15} & \textcolor{best}{73.20} \\
\bottomrule
\end{tabular}%
}
\end{table*}

\section{Experiments}

\subsection{Implementation Details}

Unless otherwise stated, we initialize the thermal branch from the MAE-pretrained checkpoint described in Appendix Sec.~\ref{sec:mae_encoder} and fine-tune \ourmodel{} for 3 epochs on a single NVIDIA H100 80GB GPU with batch size 4. Under this setup, training takes approximately 34 hours. In all experiments, we update only the last \(N_\ell=4\) blocks of the thermal encoder together with the thermal LayerNorm and the fusion module, while keeping the RGB vision tower and language model frozen.
Because Molmo's vision encoder expects 3-channel visual inputs, each single-channel thermal image is replicated across three channels before applying the standard Molmo image preprocessing pipeline. We use the same preprocessing convention at inference time and for external benchmark evaluation.

\subsection{Evaluation Protocol}

% We evaluate on Thermo-VL-Bench under three input settings: RGB-only, thermal-only, and RGB+thermal. In each setting, the model receives only the modalities required by the corresponding question. We compare \ourmodel{} against representative open-source VLM baselines using the same prompt format and input setting for each subset. We report accuracy on the human-verified QA pairs. For RGB+thermal evaluation, our model receives the aligned RGB and thermal pair directly. Baseline VLMs are given the same paired inputs in a fixed modality order using the same prompt template.

We evaluate on Thermo-VL-Bench, a binary QA benchmark with RGB-only, thermal-only, and RGB+thermal subsets. In each subset, the model receives only the modalities specified by the question. For \ourmodel{}, RGB and thermal inputs are processed jointly through the proposed fusion module. For baseline VLMs, which do not natively support RGB--thermal fusion, we use an unmodified paired-input protocol in which RGB and thermal images are presented in a fixed order under the same prompt template. These baseline results should be interpreted as out-of-the-box adaptation to paired multispectral inputs rather than fully optimized thermal-specific systems.

\subsection{Main Results on Thermo-VL-Bench}

Table~\ref{tab:rgb_thermal_results} reports performance on Thermo-VL-Bench under three input settings: RGB-only, thermal-only, and RGB+thermal. Relative to the Molmo-7B-D-0924 baseline, \ourmodel{} preserves RGB-only performance (88.67 vs.~88.67), improves thermal-only performance (82.23 vs.~81.20), and substantially improves RGB+thermal performance (88.96 vs.~75.79). The largest gain appears in the fused RGB+thermal setting, indicating that the proposed text-guided fusion module uses complementary thermal evidence rather than simply replacing the RGB pathway. Overall, \ourmodel{} achieves the best combined score among the compared models.

\subsection{Results on RGB-Th Bench}

We further evaluate \ourmodel{} on RGB-Th Bench~\cite{moshtaghi2025rgb}, an external benchmark for RGB--thermal question answering. For fair comparison, we represent thermal inputs as single-channel images and replicate the channel three times before applying the standard visual preprocessing pipeline expected by the backbone. Table~\ref{tab:rgb_th_eval_results} shows that \ourmodel{} matches the Molmo baseline in the RGB-Txt setting (77.25 vs.~77.25) while improving RGB-Th-Txt performance from 60.84 to 69.15. The gain on RGB-Th Bench is concentrated in the RGB-Th-Txt setting, which is consistent with the proposed model’s design goal of improving joint RGB--thermal grounding while preserving RGB-only behavior.

\subsection{Ablation Studies}

We ablate the proposed fusion design and training objective to isolate the effect of each component. Unless otherwise noted, all ablations use the same training data, optimization settings, and evaluation protocol as the full model. We report accuracy on Thermo-VL-Bench, and analyze how prompt-guided attention, RGB-guided attention, auxiliary regularization, and partial thermal fine-tuning contribute to performance.

\begin{table}[t]
\centering
\caption{Ablation study of fusion design and training objective on the RGB+IR subset of Thermo-VL-Bench. Accuracy is reported on the human-verified QA pairs.}
\label{tab:ablation_main}
\small
\begin{tabular}{l c | l c}
\toprule
Variant & (RGB+IR)$\uparrow$ & Variant  & (RGB+IR)$\uparrow$ \\
\midrule

\multicolumn{2}{c|}{\textit{Full model}} 
& \multicolumn{2}{c}{\textit{Loss components}} \\
Full model & 88.96 
& w/o $\mathcal{L}_{\mathrm{align}}$ & 87.61 \\
 & 
& w/o $\mathcal{L}_{\mathrm{contr}}$ & 86.30 \\
 & 
& w/o $\mathcal{L}_{\mathrm{gate}}$ & 87.94 \\

\midrule
\multicolumn{2}{c|}{\textit{Fusion architecture}} 
& \multicolumn{2}{c}{\textit{Training strategy}} \\
w/o text-guided attention & 82.34 
& w/o RGB block masking & 88.11 \\
w/o RGB-guided attention & 85.06 
& Train all thermal blocks & 86.54 \\
w/o gated residual (direct fusion) & 85.91 
& Train last $N_\ell=2$ blocks & 88.10 \\

\bottomrule
\end{tabular}
\end{table}

Table~\ref{tab:ablation_main} isolates the effects of the proposed fusion architecture and training objective. Removing text-guided attention causes the largest drop, suggesting that conditioning fusion on the question is useful. Removing RGB-guided attention or the gated residual also hurts performance, which is consistent with the goal of preserving Molmo’s pretrained RGB interface while incorporating thermal evidence. Among the auxiliary losses, \(\mathcal{L}_{\mathrm{contr}}\) contributes most to cross-modal consistency, while \(\mathcal{L}_{\mathrm{gate}}\) helps prevent collapse to a single modality. Block-wise RGB masking and partial thermal fine-tuning further improve robustness by discouraging RGB over-reliance and reducing overfitting in the thermal branch. We additionally evaluated an alternative text-conditioning strategy in which the fusion module uses frozen CLIP text embeddings instead of Molmo prompt embeddings; this variant underperformed the default design and is reported in Appendix Sec.~\ref{sec:clip_text_ablation}.

\subsection{Qualitative Analysis}

\begin{figure}[t]
  \centering
  \includegraphics[width=\columnwidth]{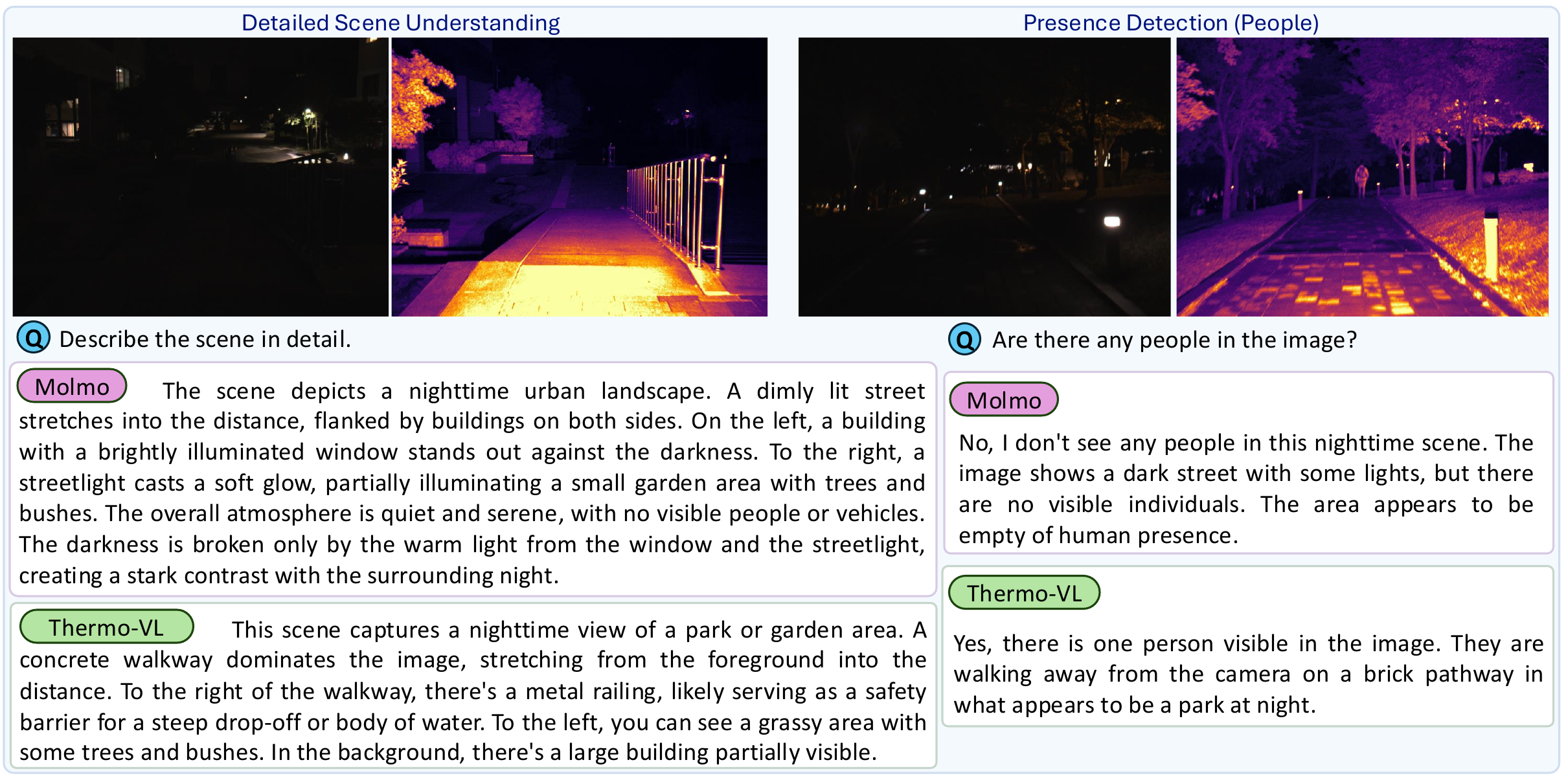}
  \caption{\textbf{Qualitative comparison in low light.} \ourmodel{} improves detailed scene understanding (left) and correctly detects a person missed by Molmo (right). Thermal images are heatmap-colorized for visualization only.}
  \label{fig:qualitative_comparison}
\end{figure}

Figure~\ref{fig:qualitative_comparison} presents two qualitative comparisons between \ourmodel{} and the RGB-only Molmo-7B-D-0924 baseline on challenging low-light scenes. In the left example, the task is detailed scene understanding. Molmo produces a plausible but generic description of a dark nighttime scene and misses structural cues. In contrast, \ourmodel{} leverages complementary thermal evidence to recover the concrete walkway, safety railing, surrounding vegetation, and broader park-like layout, yielding a more specific and spatially grounded response. In the right example, the task is presence detection. Molmo incorrectly concludes that the scene is empty, whereas \ourmodel{} correctly identifies a person visible on the pathway. 
Together, these examples highlight two benefits of the proposed fusion design. First, RGB--thermal fusion improves fine-grained scene understanding when visible-spectrum content is heavily degraded. Second, it strengthens safety-relevant presence detection by injecting thermal evidence only where it is useful for the current question. These qualitative observations are consistent with the quantitative gains reported on Thermo-VL-Bench and RGB-Th Bench.

\section*{Limitations}

% Despite the promising results, this work has several limitations. First, \ourmodel{} assumes paired and spatially aligned RGB--thermal inputs; we do not evaluate robustness to misalignment, missing modalities, or substantially different thermal sensor characteristics. Second, the training set combines real paired imagery with synthetic thermal augmentation and model-generated question--answer supervision, which may introduce artifacts, linguistic bias, or supervision noise. Finally, Thermo-VL-Bench is a manually screened binary benchmark whose candidate items are generated by a language model before human filtering; therefore, it should be viewed as an initial evaluation resource rather than a fully human-authored benchmark. For these reasons, we view \ourmodel{} as a research prototype rather than a deployment-ready system, especially for high-stakes safety-critical use.

Despite promising results, this work has several limitations. \ourmodel{} assumes paired, spatially aligned RGB--thermal inputs and does not evaluate robustness to misalignment, missing modalities, or sensor shifts. The training data combines real paired imagery with synthetic thermal augmentation and model-generated QA supervision, which may introduce artifacts or bias. Thermo-VL-Bench is a manually screened, model-generated binary benchmark and should be viewed as an initial evaluation resource rather than a fully human-authored benchmark. Accordingly, we view \ourmodel{} as a research prototype rather than a deployment-ready system, especially for safety-critical use.

\section{Conclusion}

We presented \ourmodel{}, a wavelength-aware VLM that integrates aligned RGB and thermal imagery through prompt-conditioned dual-attention fusion. By freezing the pretrained RGB--language backbone and adapting only a thermal encoder with a gated residual fusion module, \ourmodel{} incorporates complementary thermal evidence while preserving RGB performance. We also introduced an RGB--thermal instruction-tuning dataset and Thermo-VL-Bench for low-light and cross-spectrum VQA. Results on Thermo-VL-Bench and RGB-Th Bench show improved thermal and RGB+thermal reasoning, highlighting native multispectral fusion as a promising direction for robust VLMs in visually degraded settings.

% \section*{Impact Statement}

% We present a wavelength-aware multimodal framework that natively fuses thermal infrared and visible (RGB) imagery, together with the first large-scale RGB–thermal VQA dataset. This combination enables more reliable high-level reasoning in low-visibility and safety-critical settings (e.g., nighttime driving, search-and-rescue, disaster response, and infrastructure inspection) where RGB alone is unreliable.

% Finally, we emphasize this work as a research prototype intended to inform further study and careful, ethically governed deployment.

% \section*{Impact Statement}

% This work improves multimodal reasoning in low-visibility conditions by combining RGB and thermal imagery within a language-guided framework. It enables applications such as nighttime scene understanding, search-and-rescue, and infrastructure inspection, where RGB-only systems often fail. However, these capabilities may be misused for surveillance or other dual-use scenarios, and errors could create false confidence in safety-critical settings. The training data and benchmarks also inherit biases and imperfections from source datasets and model-generated annotations. We present this as a research prototype, not a stand-alone decision system, and recommend clear usage guidelines, respect for dataset licenses, and avoiding deployment in high-stakes settings without human oversight.

% \section*{Acknowledgment}


\bibliography{references}

@inproceedings{hwang2015multispectral,
  title     = {Multispectral Pedestrian Detection: Benchmark Dataset and Baseline},
  author    = {Hwang, Soonmin and Park, Jaesik and Kim, Namil and Choi, Yukyung and Kweon, In So},
  booktitle = {Proceedings of the IEEE Conference on Computer Vision and Pattern Recognition (CVPR)},
  year      = {2015}
}

@inproceedings{otcbvs_osu_thermal,
  author    = {James W. Davis and Mark A. Keck},
  title     = {A Two-Stage Approach to Person Detection in Thermal Imagery},
  booktitle = {Proceedings of the IEEE Workshop on Applications of Computer Vision (WACV)},
  year      = {2005},
  month     = {January},
  note      = {IEEE OTCBVS WS Series Bench}
}

@article{otcbvs_osu_color_thermal,
  author    = {James W. Davis and Vinay Sharma},
  title     = {Background-Subtraction using Contour-based Fusion of Thermal and Visible Imagery},
  journal   = {Computer Vision and Image Understanding},
  volume    = {106},
  number    = {2-3},
  pages     = {162--182},
  year      = {2007},
  note      = {IEEE OTCBVS WS Series Bench}
}

@InProceedings{epfl_rgb_nir_scene,
  author =   {M. Brown and S. S\"usstrunk},
  title =    {Multispectral {SIFT} for Scene Category Recognition},
  OPTcrossref =  {},
  OPTkey =   {},
  booktitle = {Computer Vision and Pattern Recognition (CVPR11)},
  pages =     {177--184},
  year =     {2011},
  OPTeditor =    {},
  OPTvolume =    {},
  OPTnumber =    {},
  OPTseries =    {},
  address =      {Colorado Springs},
  month =    {June},
  OPTorganization = {},
  OPTpublisher = {},
  OPTnote =     {},
  OPTannote =    {}
}

@article{moshtaghi2025rgb,
  title={RGB-Th-Bench: A Dense benchmark for Visual-Thermal Understanding of Vision Language Models},
  author={Moshtaghi, Mehdi and Khajavi, Siavash H and Pajarinen, Joni},
  journal={arXiv preprint arXiv:2503.19654},
  year={2025}
}

@inproceedings{deitke2025molmo,
  title={Molmo and pixmo: Open weights and open data for state-of-the-art vision-language models},
  author={Deitke, Matt and Clark, Christopher and Lee, Sangho and Tripathi, Rohun and Yang, Yue and Park, Jae Sung and Salehi, Mohammadreza and Muennighoff, Niklas and Lo, Kyle and Soldaini, Luca and others},
  booktitle={Proceedings of the Computer Vision and Pattern Recognition Conference},
  pages={91--104},
  year={2025}
}

@inproceedings{deitke2025molmo1,
  title={Molmo and pixmo: Open weights and open data for state-of-the-art vision-language models},
  author={Deitke, Matt and Clark, Christopher and Lee, Sangho and Tripathi, Rohun and Yang, Yue and Park, Jae Sung and Salehi, Mohammadreza and Muennighoff, Niklas and Lo, Kyle and Soldaini, Luca and others},
  booktitle={Proceedings of the Computer Vision and Pattern Recognition Conference},
  pages={91--104},
  year={2025}
}

@article{huang2025visual,
  title={Visual instruction tuning towards general-purpose multimodal large language model: A survey},
  author={Huang, Jiaxing and Zhang, Jingyi and Jiang, Kai and Qiu, Han and Zhang, Xiaoqin and Shao, Ling and Lu, Shijian and Tao, Dacheng},
  journal={International Journal of Computer Vision},
  volume={133},
  number={11},
  pages={8151--8189},
  year={2025},
  publisher={Springer}
}

@article{liu2023visual,
  title={Visual instruction tuning},
  author={Liu, Haotian and Li, Chunyuan and Wu, Qingyang and Lee, Yong Jae},
  journal={Advances in neural information processing systems},
  volume={36},
  pages={34892--34916},
  year={2023}
}

@misc{rgb_nir_scene_dataset,
  title = {RGB-NIR Scene Dataset},
  author = {Brown, Michael and S{\"u}sstrunk, Sabine},
  year = {2011},
  howpublished = {\url{https://www.epfl.ch/labs/ivrl/research/downloads/rgb-nir-scene-dataset/}},
  note = {IVRL Lab, EPFL}
}

@article{zhao2024removal,
  title={Removal and selection: Improving rgb-infrared object detection via coarse-to-fine fusion},
  author={Zhao, Tianyi and Yuan, Maoxun and Jiang, Feng and Wang, Nan and Wei, Xingxing},
  journal={arXiv preprint arXiv:2401.10731},
  year={2024}
}

@article{paranjape2025f,
  title={F-ViTA: Foundation Model Guided Visible to Thermal Translation},
  author={Paranjape, Jay N and de Melo, Celso and Patel, Vishal M},
  journal={arXiv preprint arXiv:2504.02801},
  year={2025}
}

@inproceedings{shyam2024lightweight,
  title={Lightweight thermal super-resolution and object detection for robust perception in adverse weather conditions},
  author={Shyam, Pranjay and Yoo, HyunJin},
  booktitle={Proceedings of the IEEE/CVF Winter Conference on Applications of Computer Vision},
  pages={7471--7482},
  year={2024}
}

@inproceedings{jia2021llvip,
  title={LLVIP: A Visible-infrared Paired Dataset for Low-light Vision},
  author={Jia, Xinyu and Zhu, Chuang and Li, Minzhen and Tang, Wenqi and Zhou, Wenli},
  booktitle={Proceedings of the IEEE/CVF International Conference on Computer Vision},
  pages={3496--3504},
  year={2021}
}

@dataset{flir_adas,
  author       = {{Teledyne FLIR}},
  title        = {Teledyne {FLIR} Thermal Dataset for Algorithm Training},
  year         = {2018},
  publisher    = {FLIR Systems, Inc.},
  howpublished = {\url{https://oem.flir.com/solutions/automotive/dataset/}},
  note         = {Thermal and visible spectrum frames for ADAS and autonomous vehicle development}
}

@inproceedings{jiang2024infraredllava,
  title={Infrared-LLaVA: Enhancing Understanding of Infrared Images in Multi-Modal Large Language Models},
  author={Jiang, Shixin and Chen, Zerui and Liang, Jiafeng and Zhao, Yanyan and Liu, Ming and Qin, Bing},
  booktitle={Findings of the Association for Computational Linguistics: EMNLP 2024},
  pages={8573--8591},
  year={2024},
  address={Miami, Florida, USA},
  publisher={Association for Computational Linguistics}
}

@inproceedings{el2023enhanced,
  title={Enhanced Thermal-RGB Fusion for Robust Object Detection},
  author={El Ahmar, Wassim and others},
  booktitle={Proceedings of the IEEE/CVF Conference on Computer Vision and Pattern Recognition Workshops},
  pages={2312--2321},
  year={2023}
}

@article{sun2019rtfnet,
  title={RTFNet: RGB-Thermal Fusion Network for Semantic Segmentation of Urban Scenes},
  author={Sun, Yuxiang and Zuo, Weixun and Liu, Ming},
  journal={IEEE Robotics and Automation Letters},
  volume={4},
  number={3},
  pages={2576--2583},
  year={2019},
  publisher={IEEE}
}

@article{xie2023illumination,
  title={Illumination-guided with Crossmodal Transformer Fusion for RGB-T Object Detection},
  author={Xie, Ruilin and Jiang, Sheng and Bie, Yiming and Xia, Miaolei},
  journal={Journal of Imaging Science and Technology},
  volume={69},
  number={2},
  pages={020505--1},
  year={2023},
  doi={10.2352/J.ImagingSci.Technol.2025.69.2.020505}
}

@misc{wu2025atdetr,
  title={AT-Detr: A Multispectral Detector with Adaptive Feature Alignment and Fusion},
  author={Wu, H and others},
  year={2025},
  note={\url{https://papers.ssrn.com/sol3/papers.cfm?abstract_id=5099705}}
}

@article{li2025multimodal,
  title={Multimodal fusion transformer network for multispectral pedestrian detection in low-light condition},
  author={Li, Gong and Ren, Guoyin and Wang, Jingyu and Zhi, Mobing and Yu, Zhijie and Jiang, Bo and Guan, Haoliang and Guo, Qidan},
  journal={Scientific Reports},
  volume={15},
  number={1},
  pages={18778},
  year={2025},
  publisher={Nature Publishing Group UK London}
}

@article{rathinam2024hybrid,
  title={Hybrid Attention for Robust RGB-T Pedestrian Detection in Real-World Conditions},
  author={Rathinam, Arunkumar and Pauly, Leo and Rharbaoui, Wassim and Kacem, Anis and Gaudilli{\`e}re, Vincent and Aouada, Djamila and others},
  journal={IEEE Robotics and Automation Letters},
  year={2024},
  publisher={IEEE}
}

@article{xiao2025thermalgen,
  author = {Xiao, Jiuhong and Nayak, Roshan and Zhang, Ning and Tortei, Daniel and Loianno, Giuseppe},
  title = {ThermalGen: Style-Disentangled Flow-Based Generative Models for RGB-to-Thermal Image Translation},
  journal = {arXiv preprint arXiv:2509.24878},
  year = {2025},
  note = {Accepted at NeurIPS 2025},
  doi = {10.48550/arXiv.2509.24878},
  url = {https://arxiv.org/abs/2509.24878}
}

@inproceedings{berg2018generating,
  author = {Berg, Andreas and Olofsson, Mikael and Johanson, Fredrik},
  title = {Generating Visible Spectrum Images from Thermal Infrared},
  booktitle = {Proceedings of the IEEE Conference on Computer Vision and Pattern Recognition Workshops (CVPRW)},
  year = {2018},
  url = {https://openaccess.thecvf.com/content_cvpr_2018_workshops/w21/html/Berg_Generating_Visible_Spectrum_CVPR_2018_paper.html}
}

@article{li2023feasibility,
  author = {Li, Yuchuan and Ko, Yoon and Lee, Wonsook},
  title = {A Feasibility Study on Translation of RGB Images to Thermal Images: Development of a Machine Learning Algorithm},
  journal = {SN Computer Science},
  volume = {4},
  number = {5},
  year = {2023},
  doi = {10.1007/s42979-023-02040-4},
  url = {https://dl.acm.org/doi/10.1007/s42979-023-02040-4}
}

@inproceedings{chen2024internvl,
  title={Internvl: Scaling up vision foundation models and aligning for generic visual-linguistic tasks},
  author={Chen, Zhe and Wu, Jiannan and Wang, Wenhai and Su, Weijie and Chen, Guo and Xing, Sen and Zhong, Muyan and Zhang, Qinglong and Zhu, Xizhou and Lu, Lewei and others},
  booktitle={Proceedings of the IEEE/CVF conference on computer vision and pattern recognition},
  pages={24185--24198},
  year={2024}
}

@article{Qwen-VL,
  title={Qwen-VL: A Versatile Vision-Language Model for Understanding, Localization, Text Reading, and Beyond},
  author={Bai, Jinze and Bai, Shuai and Yang, Shusheng and Wang, Shijie and Tan, Sinan and Wang, Peng and Lin, Junyang and Zhou, Chang and Zhou, Jingren},
  journal={arXiv preprint arXiv:2308.12966},
  year={2023}
}

@misc{liu2024llavanext,
    title={LLaVA-NeXT: Improved reasoning, OCR, and world knowledge},
    url={https://llava-vl.github.io/blog/2024-01-30-llava-next/},
    author={Liu, Haotian and Li, Chunyuan and Li, Yuheng and Li, Bo and Zhang, Yuanhan and Shen, Sheng and Lee, Yong Jae},
    month={January},
    year={2024}
}

@article{o2015introduction,
  title={An introduction to convolutional neural networks},
  author={O'shea, Keiron and Nash, Ryan},
  journal={arXiv preprint arXiv:1511.08458},
  year={2015}
}

@inproceedings{radford2021learning,
  title={Learning transferable visual models from natural language supervision},
  author={Radford, Alec and Kim, Jong Wook and Hallacy, Chris and Ramesh, Aditya and Goh, Gabriel and Agarwal, Sandhini and Sastry, Girish and Askell, Amanda and Mishkin, Pamela and Clark, Jack and others},
  booktitle={International conference on machine learning},
  pages={8748--8763},
  year={2021},
  organization={PmLR}
}
\bibliographystyle{plainnat}

%%%%%%%%%%%%%%%%%%%%%%%%%%%%%%%%%%%%%%%%%%%%%%%%%%%%%%%%%%%%

\clearpage
\appendix

\section{Technical appendices and supplementary material}

\subsection{MAE encoder training}
\label{sec:mae_encoder}

\begin{figure}[!h]
    \centering
    \includegraphics[width=.65\linewidth]{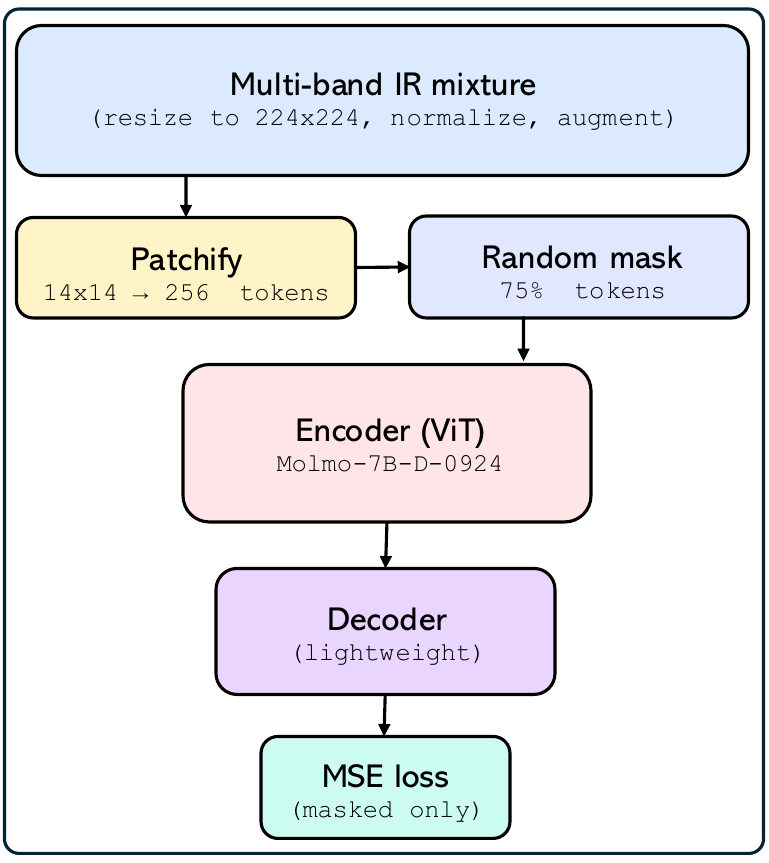}
    \caption{MAE encoder pre-training on a pooled infrared corpus (KAIST, FLIR-ADAS, OSU/OTCBVS, NIRSCENE). Inputs are resized to \(224\times224\), patchified into 256 tokens, 75\% randomly masked, encoded by a ViT backbone, and reconstructed by a lightweight decoder. MSE loss is applied on masked patches.}
    \label{fig:mae_encoder_pipeline}
\end{figure}

To obtain a robust thermal feature extractor, we pre-train a Masked Autoencoder (MAE) on a pooled multi-band infrared corpus (See figure ~\ref{fig:mae_encoder_pipeline}) built from four public datasets: KAIST, FLIR-ADAS, OSU (OTCBVS), and NIRSCENE. KAIST provides aligned color--thermal pairs for pedestrian-centric driving scenes and explicitly characterizes thermal sensing in the long-wave infrared (LWIR) band (thermal band, 7.5--13~\(\mu\)m), while also contrasting it with near-infrared (NIR, 0.75--1.3~\(\mu\)m) imaging \cite{hwang2015multispectral}. FLIR-ADAS contributes large-scale annotated thermal/visible automotive data \cite{flir_adas}. The OSU OTCBVS benchmark adds additional thermal imagery and RGB--thermal collections recorded in outdoor pedestrian scenes \cite{otcbvs_osu_thermal,otcbvs_osu_color_thermal}. Finally, NIRSCENE (RGB--NIR) broadens the training distribution with near-infrared scene content \cite{rgb_nir_scene_dataset}. Training on this mixture encourages the encoder to learn modality-invariant structures (e.g., object boundaries and coarse geometry) that transfer across infrared bands and remain stable across diverse imaging conditions such as illumination, temperature variation, and sensor characteristics.

\paragraph{MAE pre-training objective}
We follow the asymmetric MAE design in which the encoder operates only on unmasked patches and a lightweight decoder reconstructs masked content in pixel space \cite{he2022mae}. Each input image is resized to \(224\times224\) and partitioned into non-overlapping \(14\times14\) patches, producing \(16\times16=256\) tokens. We randomly mask 75\% of tokens per image and feed the remaining 25\% to the encoder backbone (Molmo-7B-D-0924 ViT). The decoder receives the encoded visible tokens plus learned mask tokens (restored to original order using positional embeddings) and predicts all patch pixels. The reconstruction loss is computed only over masked patches:
\begin{equation}
\mathcal{L}_{\mathrm{MAE}}
= \frac{1}{|\mathcal{M}|}\sum_{i\in\mathcal{M}}
\left\lVert \mathbf{x}_i - \hat{\mathbf{x}}_i \right\rVert_2^2,
\end{equation}
where \(\mathcal{M}\) is the set of masked patch indices and \(\mathbf{x}_i\), \(\hat{\mathbf{x}}_i\) are the ground-truth and reconstructed patch vectors, respectively \cite{he2022mae}. After pre-training, the decoder is discarded and the encoder is used to initialize the thermal feature extractor.

\subsection{Additional ablation: alternative text encoder}
\label{sec:clip_text_ablation}

We additionally tested whether the fusion module benefits from external text representations by replacing Molmo prompt embeddings with features from a frozen CLIP text encoder. Concretely, we replace the prompt features used by the fusion module with CLIP text features followed by a learned projection into the visual space, while keeping the rest of the architecture unchanged. Table~\ref{tab:clip_text_ablation} shows that this variant performs substantially worse than using native Molmo prompt embeddings.

\begin{table}[h]
\centering
\caption{Alternative text-conditioning ablation on Thermo-VL-Bench (RGB+IR). Replacing native Molmo prompt embeddings with frozen CLIP text features degrades performance.}
\label{tab:clip_text_ablation}
\small
\begin{tabular}{lc}
\toprule
Text-conditioning strategy & RGB+IR $\uparrow$ \\
\midrule
Molmo prompt embeddings (ours) & 88.96 \\
Frozen CLIP text encoder & 83.10 \\
\bottomrule
\end{tabular}
\end{table}

We hypothesize that the drop arises from a representation mismatch: although CLIP text features provide a strong standalone text space, they are not naturally aligned with the token space used by the frozen Molmo language model. As a result, question-conditioned fusion is weaker when CLIP text features replace the model's native prompt embeddings.

\subsection{Algorithms for Training-VLM and Text-Guided Dual-Attention}

\[
\begin{array}{l}
\textbf{Algorithm 1: Training Frozen-VLM Thermal Fusion with Text-Guided Dual-Attention} \\[2pt]
\hline
\textbf{Input: } 
(I^{\mathrm{rgb}}, I^{\mathrm{th}}, p, y^\star),\
\phi_R,\ g_{\mathrm{LM}},\ \phi_T,\ \mathrm{LN}_T,\ \mathcal{F}_\theta \\[2pt]

\textbf{Frozen modules: }
\phi_R \text{ (RGB ViT encoder)},\ 
g_{\mathrm{LM}} \text{ (LLM head)} \\[2pt]

\textbf{Trainable modules: }
\phi_T \text{ (last-}N_\ell\text{ blocks only)},\ 
\mathrm{LN}_T,\ 
\mathcal{F}_\theta \\[2pt]

\textbf{Hyperparameters: }
\eta_{\mathrm{therm}},\ \eta_{\mathrm{fusion}},\ \mathrm{wd},\ \rho,\ \tau,\ 
\lambda_{\mathrm{align}},\ \lambda_{\mathrm{contr}},\ \lambda_{\mathrm{gate}} \\[2pt]

\textbf{Output: } 
(\phi_T,\ \mathrm{LN}_T,\ \mathcal{F}_\theta) \\[2pt]
\hline
1:\quad \tilde I^{\mathrm{rgb}} \leftarrow \mathrm{BlockMask}_{\rho}(I^{\mathrm{rgb}})
\qquad \text{\# apply contiguous RGB block masking during training} \\[2pt]

2:\quad \mathbf{R} \leftarrow \phi_R(\tilde I^{\mathrm{rgb}})
\qquad \text{\# frozen RGB patch tokens from masked RGB input} \\[2pt]

3:\quad \mathbf{T} \leftarrow \mathrm{LN}_T\!\bigl(\phi_T(I^{\mathrm{th}})\bigr)
\qquad \text{\# thermal patch tokens} \\[2pt]

4:\quad \mathbf{P} \leftarrow \mathrm{Embed}_{\mathrm{LLM}}(p)
\qquad \text{\# prompt/token embeddings} \\[2pt]

5:\quad (\mathbf{R}^{\dagger}, \boldsymbol{\alpha}) \leftarrow
\mathcal{F}_\theta(\mathbf{R}, \mathbf{T}, \mathbf{P})
\qquad \text{\# fusion module; see Algorithm 2} \\[2pt]

6:\quad \hat{y} \leftarrow g_{\mathrm{LM}}(\mathbf{R}^{\dagger}, \mathbf{P})
\qquad \text{\# frozen VLM forward} \\[2pt]

7:\quad \mathcal{L}_{\mathrm{LM}} \leftarrow \operatorname{CE}(\hat{y}, y^\star)
\qquad \text{\# main task loss} \\[2pt]

8:\quad \mathcal{L}_{\mathrm{align}} \leftarrow
\tfrac{1}{N}\sum_{n=1}^{N}\lVert \mathbf{R}_{n}-\mathbf{T}_{n}\rVert_2^2
\qquad \text{\# token-level RGB--thermal alignment} \\[2pt]

9:\quad \mathbf{p}_R \leftarrow \operatorname{Pool}(\mathbf{R}^{\dagger}),\quad
\mathbf{p}_T \leftarrow \operatorname{Pool}(\mathbf{T}),\quad
\mathbf{p}_P \leftarrow \operatorname{Pool}_{M_P}(\mathbf{P}) \\[2pt]

10:\quad \mathcal{L}_{\mathrm{contr}} \leftarrow
\tfrac{1}{2}\Bigl[
\operatorname{NCE}(\mathbf{p}_R,\mathbf{p}_T;\tau)
+
\operatorname{NCE}(\mathbf{p}_R,\mathbf{p}_P;\tau)
\Bigr]
\qquad \text{\# contrastive regularization} \\[2pt]

11:\quad H(\boldsymbol{\alpha}) \leftarrow
-\boldsymbol{\alpha}\log\boldsymbol{\alpha}
-(1-\boldsymbol{\alpha})\log(1-\boldsymbol{\alpha}) \\[2pt]

12:\quad \mathcal{L}_{\mathrm{gate}} \leftarrow
-\tfrac{1}{N}\sum_{n=1}^{N} H(\alpha_n)
\qquad \text{\# discourages saturated token gates} \\[2pt]

13:\quad \mathcal{L} \leftarrow
\mathcal{L}_{\mathrm{LM}}
+ \lambda_{\mathrm{align}}\mathcal{L}_{\mathrm{align}}
+ \lambda_{\mathrm{contr}}\mathcal{L}_{\mathrm{contr}}
+ \lambda_{\mathrm{gate}}\mathcal{L}_{\mathrm{gate}} \\[2pt]

14:\quad \text{Update Group-T: last-}N_\ell\text{ blocks of }\phi_T
\text{ with AdamW}(\eta_{\mathrm{therm}}, \mathrm{wd}) \\[2pt]

15:\quad \text{Update Group-F: }
\{\mathcal{F}_\theta,\mathrm{LN}_T\}
\text{ with AdamW}(\eta_{\mathrm{fusion}}, \mathrm{wd})
\qquad \text{\# dual-group optimization} \\[2pt]

16:\quad \textbf{return } (\phi_T,\ \mathrm{LN}_T,\ \mathcal{F}_\theta) \\[2pt]
\hline
\end{array}
\]

\[
\begin{array}{l}
\textbf{Algorithm 2: Text-Guided Dual-Attention Fusion and Gating Block} \\[2pt]
\hline
\textbf{Input: } 
\mathbf{R}\in\mathbb{R}^{N\times d},\ 
\mathbf{T}\in\mathbb{R}^{N\times d},\ 
\mathbf{P}\in\mathbb{R}^{L\times d_p},\ 
M_P \\[2pt]

\textbf{Trainable parameters: } 
W_P \text{ (text-to-visual projection)},\
\operatorname{MLP}_m \text{ (multisource merger)},\\ \
\operatorname{MLP}_r \text{ (RGB residual predictor)},\
\operatorname{MLP}_g \text{ (gate predictor)} \\[2pt]

\textbf{Output: } 
\mathbf{R}^{\dagger}\in\mathbb{R}^{N\times d}\ \text{(fused RGB tokens)},\ 
\boldsymbol{\alpha}\in(0,1)^{N\times 1}\ \text{(token-wise gates)} \\[2pt]
\hline
1:\quad \bar{\mathbf{R}} \leftarrow \operatorname{LN}(\mathbf{R}),\quad
\bar{\mathbf{T}} \leftarrow \operatorname{LN}(\mathbf{T}),\quad
\bar{\mathbf{P}} \leftarrow \operatorname{LN}(W_P\mathbf{P})
\qquad \text{\# project text into visual feature space} \\[2pt]

2:\quad \mathbf{T}^{\mathrm{txt}} \leftarrow
\operatorname{MHA}(\bar{\mathbf{T}}, \bar{\mathbf{P}}, \bar{\mathbf{P}}; M_P)
\qquad \text{\# text-guided thermal attention} \\[2pt]

3:\quad \mathbf{T}^{\mathrm{rgb}} \leftarrow
\operatorname{MHA}(\bar{\mathbf{T}}, \bar{\mathbf{R}}, \bar{\mathbf{R}})
\qquad \text{\# RGB-guided thermal attention} \\[2pt]

4:\quad \hat{\mathbf{T}} \leftarrow
\operatorname{LN}\!\Bigl(
\mathbf{T} +
\operatorname{MLP}_m\bigl(
[\bar{\mathbf{T}};\,\mathbf{T}^{\mathrm{txt}};\,\mathbf{T}^{\mathrm{rgb}}]
\bigr)
\Bigr)
\qquad \text{\# merged thermal evidence} \\[2pt]

5:\quad \Delta\mathbf{R} \leftarrow
\operatorname{MLP}_r\!\bigl(\operatorname{LN}(\hat{\mathbf{T}})\bigr)
\qquad \text{\# candidate residual injected into RGB tokens} \\[2pt]

6:\quad \mathbf{z} \leftarrow
\operatorname{MLP}_g\!\bigl(\operatorname{LN}([\bar{\mathbf{R}};\,\hat{\mathbf{T}}])\bigr)
\qquad \text{\# gate logits from RGB/thermal context} \\[2pt]

7:\quad \boldsymbol{\alpha} \leftarrow \sigma(\mathbf{z})
\qquad \text{\# token-wise soft gates} \\[2pt]

8:\quad \mathbf{R}^{\dagger} \leftarrow
\mathbf{R} + \boldsymbol{\alpha}\odot\Delta\mathbf{R}
\qquad \text{\# gated residual fusion} \\[2pt]

9:\quad \textbf{return } \mathbf{R}^{\dagger},\ \boldsymbol{\alpha} \\[2pt]
\hline
\end{array}
\]

\subsection{Data provenance and licensing}

Our images are sourced from public datasets and remain subject to the original licenses and terms of use. We do not claim ownership of the underlying images; attribution and licensing remain with the original dataset authors. As with the source datasets, the collected data may inherit annotation noise, sampling biases, and privacy-related limitations documented by the original releases.

\paragraph{Generation of questions and answers.}
Questions and reference answers are generated using OpenAI models. As a result, the generated text may contain residual hallucinations, omissions, or social biases. We apply filtering and, where noted, human verification, but the resulting dataset should not be treated as infallible ground truth.

\paragraph{Responsible use.}
Because the dataset combines reused imagery with model-generated language, downstream users should validate outputs before high-stakes deployment. Models trained on this data may reproduce artifacts from both the source datasets and the generation models.

\subsection{Documentation for released assets}

\paragraph{Released assets.}
This work introduces two new assets: (i) a pixel-aligned RGB--thermal instruction-tuning dataset with question--answer pairs, and (ii) Thermo-VL-Bench, a human-verified RGB--thermal evaluation benchmark. We also plan to release training and evaluation scripts, model configuration files, and instructions for reproducing the reported experiments.

\paragraph{Intended use.}
The new assets are intended for research on multispectral vision--language learning, instruction tuning, and benchmark evaluation under low-light and cross-spectrum conditions. They are not intended to serve as a stand-alone basis for operational or safety-critical decision making.

\subsection{Question answer pair generation}

\begin{figure}[!h]
  \centering
  \includegraphics[width=\columnwidth]{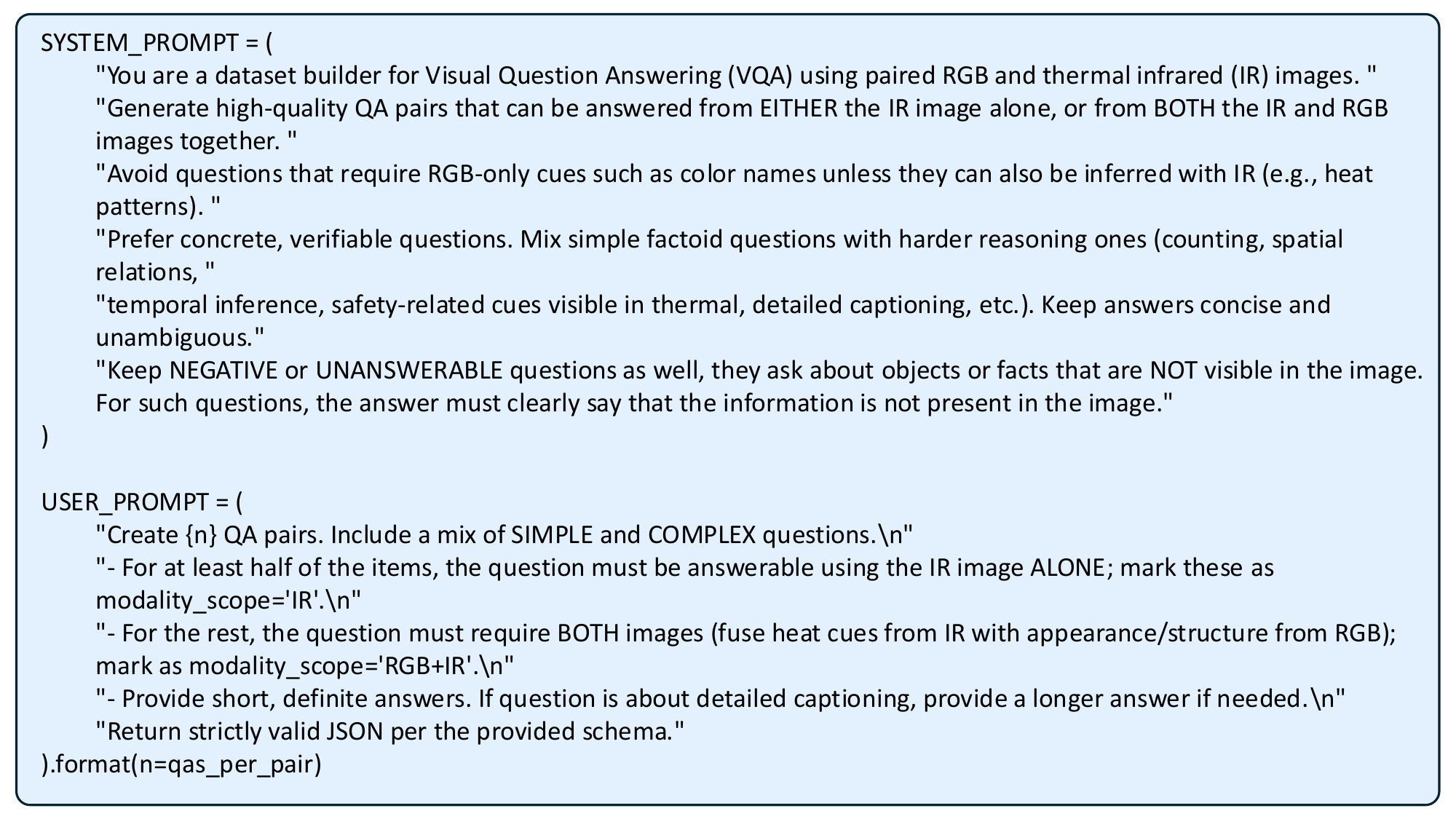}
  \caption{Prompt used for QA generation for train data}
  \label{fig:train_qa_generation_prompt}
\end{figure}

Figure \ref{fig:train_qa_generation_prompt} shows the prompt used with OpenAI GPT-5-mini to generate QA pairs for the training data. This prompt is applied only to a subset of the dataset, since part of the data is taken from the filtered PixMoCap-QA corpus, which already contains QA pairs.

\begin{figure}[!h]
  \centering
  \includegraphics[width=\columnwidth]{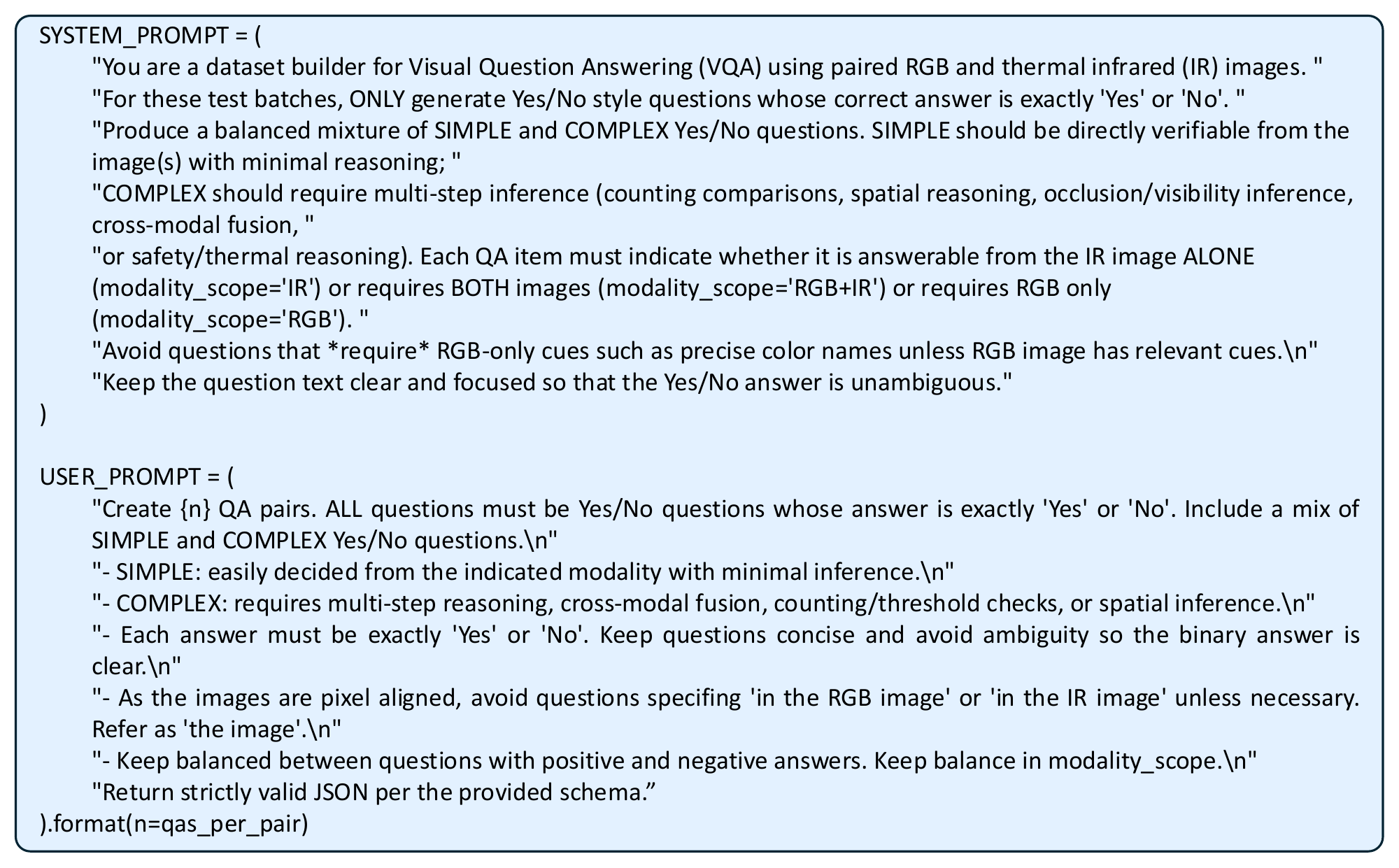}
  \caption{Prompt used for QA generation for evaluation benchmark}
  \label{fig:eval_qa_generation_prompt}
\end{figure}

Figure~\ref{fig:eval_qa_generation_prompt} shows the prompt used with OpenAI GPT-5.1 to generate QA pairs for the evaluation benchmark.

\subsection{Data Statistics}

\subsubsection{Train Dataset Statistics}

\begin{figure}[!h]
  \centering
  \includegraphics[width=\columnwidth]{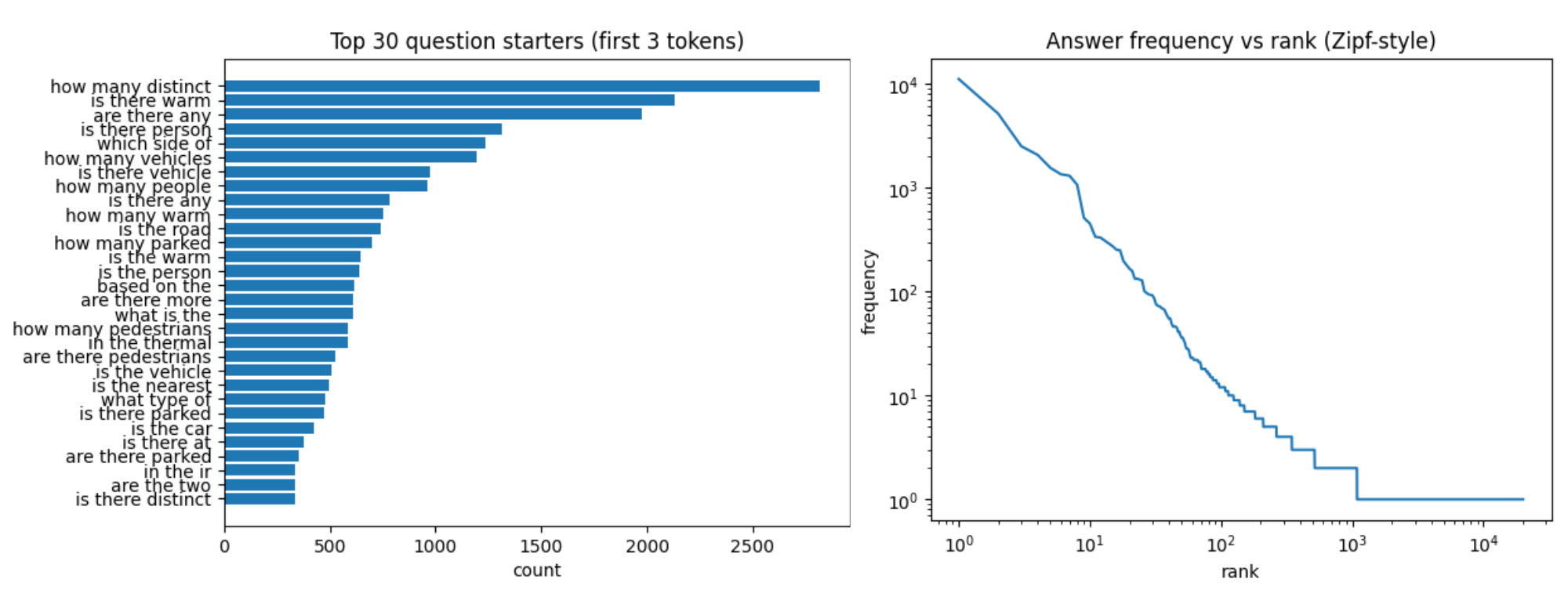}
  \caption{Data statistics for the training dataset. Left: top 30 question starters for the first 3 tokens. Right: answer frequency vs. rank.}
  \label{fig:train-data-stats}
\end{figure}

Figure~\ref{fig:train-data-stats} provides an overview of the training dataset.
The left panel reports the most common question starters (first three tokens), showing that prompts are organized around a set of recurring linguistic patterns that reflect the dataset’s task structure.
The right panel plots answer frequency versus rank and highlights a characteristic head--tail profile: a compact set of high-frequency answers is complemented by a broad set of less frequent ones.
Together, these statistics indicate that the training set contains both strong coverage of common cases and diverse, lower-frequency examples, which we account for in our training.

\subsubsection{Evaluation data distribution}
\label{sec:eval_data_distribution}

Table~\ref{tab:eval-data-distribution} summarizes the distribution of QA pairs categorized by modality, including RGB-only, Thermal-only, and combined RGB+Thermal questions.

\begin{table}[h]
\centering
\caption{Distribution of question--answer (QA) pairs across modalities in the evaluation benchmark.}
\tiny
\resizebox{.6\columnwidth}{!}{%
\begin{tabular}{c|c|c|c}
\hline
\textbf{Answer Type} & \textbf{RGB} & \textbf{IR} & \textbf{RGB+IR} \\
\hline
YES & 42.60\% & 48.90\% & 50.12\% \\
\hline
NO & 57.40\% & 51.10\% & 49.88\% \\
\hline
\end{tabular}%
}
\label{tab:eval-data-distribution}
\end{table}

\subsection{Additional Data Visualizations}

\subsubsection{Training data samples}

\begin{figure*}[!h]
    \centering
    \includegraphics[width=\textwidth]{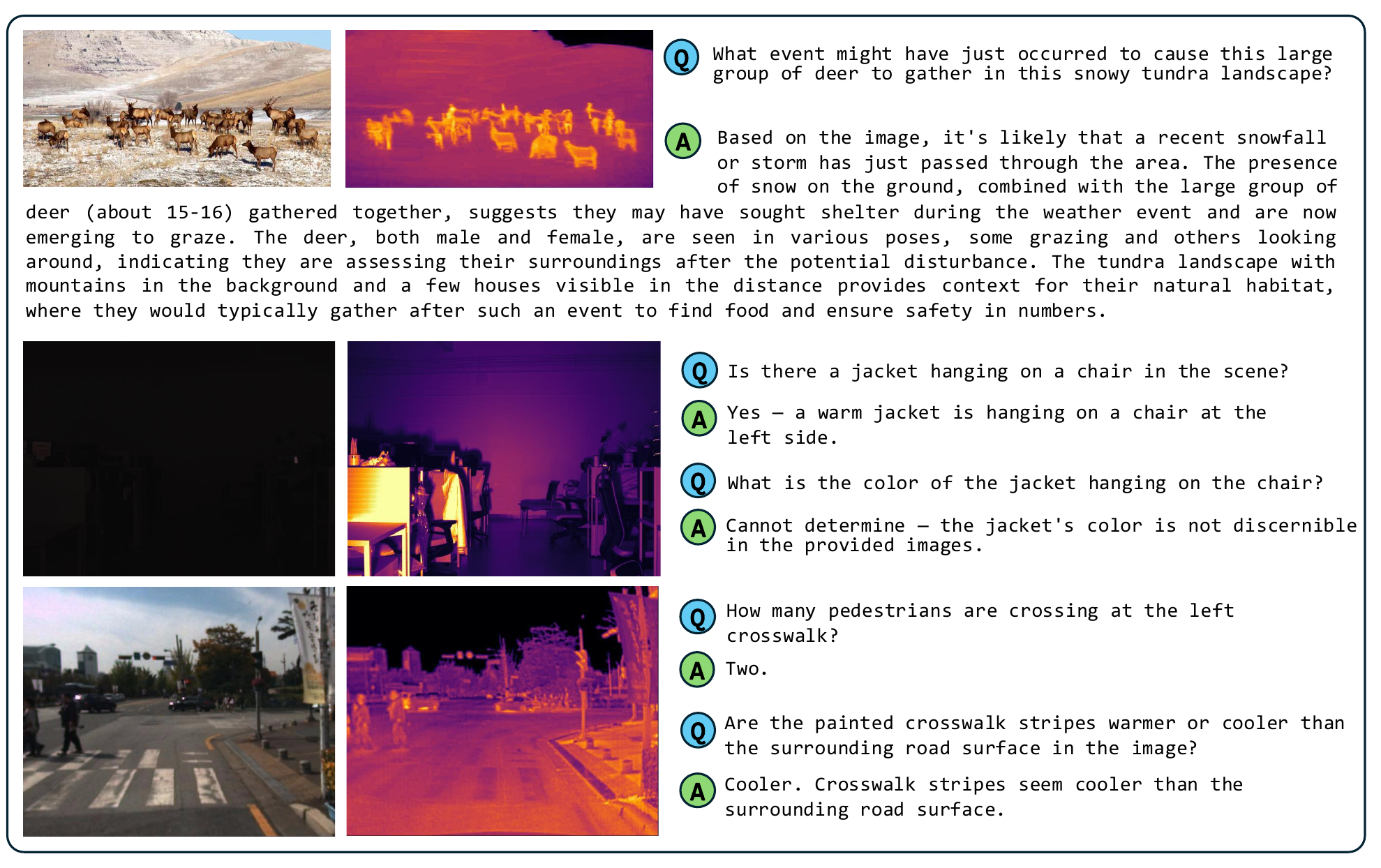}
    \caption{
        Additional samples from the training dataset. 
        Each sample includes an RGB image (left), a thermal image (right), and corresponding question--answer pairs.
        Thermal images are displayed using heatmap colorization to enhance visual interpretability.
    }
    \label{fig:more_train_data_samples}
\end{figure*}

Figure \ref{fig:more_train_data_samples} shows representative examples from the training split, illustrating paired RGB (left) and heatmap-colorized thermal images (right) alongside the associated question–answer annotations.

\begin{figure*}[!h]
    \centering
    \includegraphics[width=\textwidth]{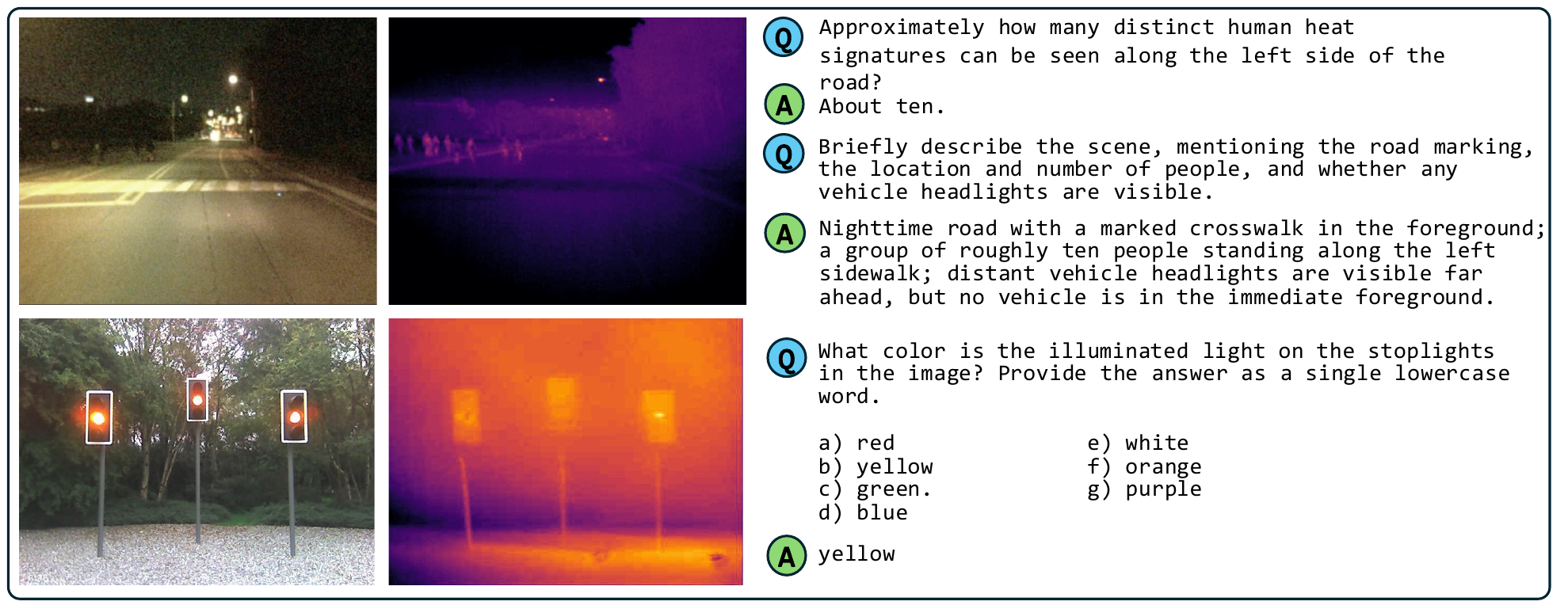}
    \caption{
    Representative samples from the training dataset.
    Each sample includes an RGB image (left), a single-channel thermal image visualized with a heatmap (right), and corresponding question--answer pairs.
    The heatmap colorization is for visualization only; model inputs use single-channel thermal images replicated to 3 channels.
    }
    \label{fig:train_data_sample}
\end{figure*}

\subsubsection{Evaluation data samples}

\begin{figure}[!t]
    \centering
    \includegraphics[width=\textwidth]{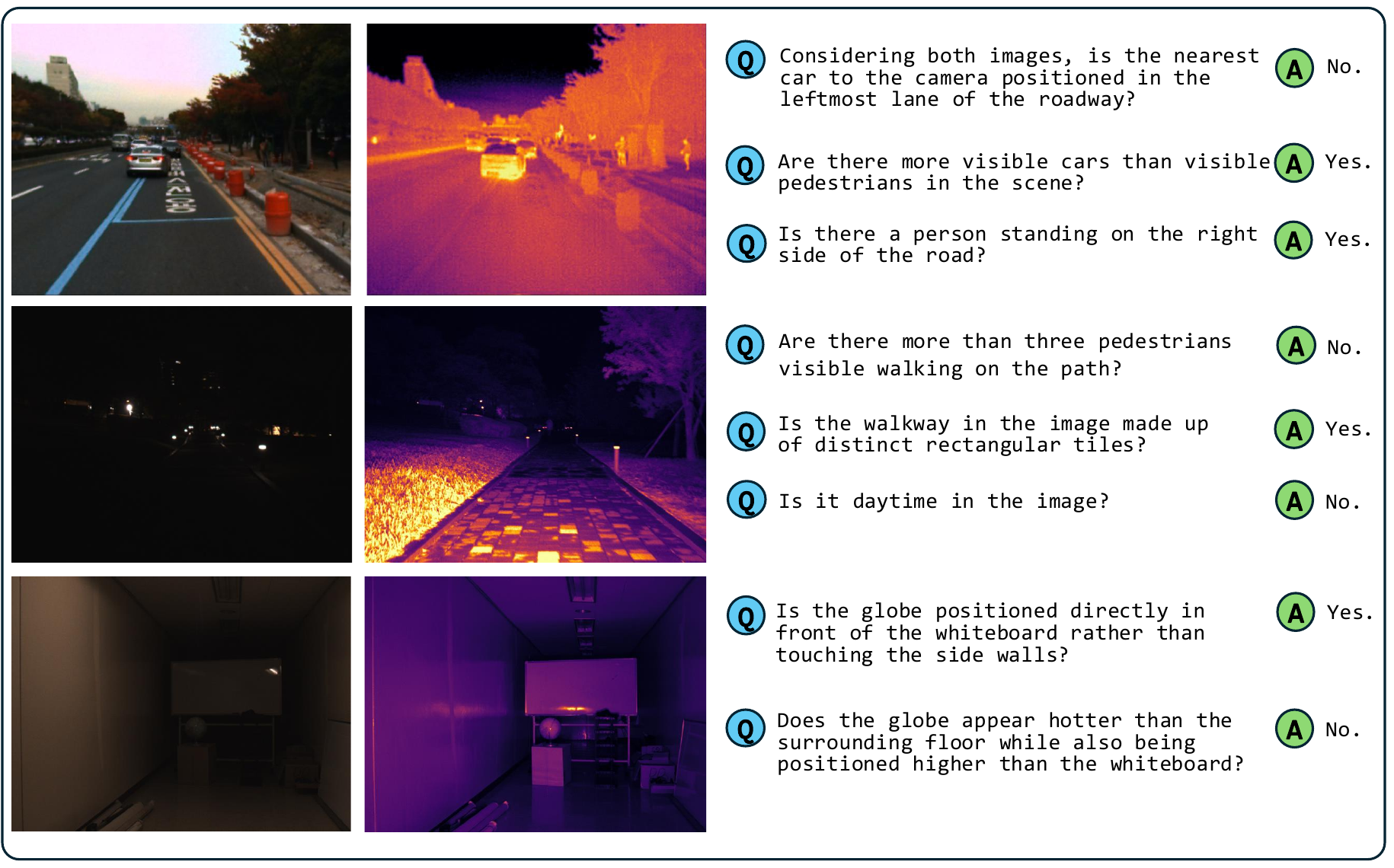}
    \caption{
        Representative samples from the evaluation dataset. 
        Each sample includes an RGB image (left), a thermal image (right), and corresponding question--answer pairs.
        Thermal images are displayed using heatmap colorization to enhance visual interpretability.
    }
    \label{fig:more_evaluation_data_samples}
\end{figure}

Figure \ref{fig:more_evaluation_data_samples} presents representative evaluation examples in the same format, highlighting how the model is assessed on RGB–thermal pairs with corresponding QA pairs.

\subsection{Broader Impact}

This work improves multimodal reasoning in low-visibility conditions by combining RGB and thermal imagery within a language-guided framework. It enables applications such as nighttime scene understanding, search-and-rescue, and infrastructure inspection, where RGB-only systems often fail. However, these capabilities may be misused for surveillance or other dual-use scenarios, and errors could create false confidence in safety-critical settings. 

The training data and benchmarks also inherit biases and imperfections from source datasets and model-generated annotations. We present this as a research prototype, not a stand-alone decision system, and recommend clear usage guidelines, respect for dataset licenses, and avoiding deployment in high-stakes settings without human oversight.

\clearpage
%%%%%%%%%%%%%%%%%%%%%%%%%%%%%%%%%%%%%%%%%%%%%%%%%%%%%%%%%%%%

% \newpage
% \input{checklist.tex}

\end{document}